\pdfoutput=1
  
\documentclass{article} %
\usepackage{conference,times}

\usepackage{amsmath,amsfonts,bm}

\def\figref#1{Figure~\ref{#1}}
\def\Figref#1{Figure~\ref{#1}}

\def\secref#1{Section~\ref{#1}}

\def\eqref#1{Equation~\ref{#1}}

\def\1{\bm{1}}

\DeclareMathAlphabet{\mathsfit}{\encodingdefault}{\sfdefault}{m}{sl}
\SetMathAlphabet{\mathsfit}{bold}{\encodingdefault}{\sfdefault}{bx}{n}

\usepackage{url}
\usepackage{graphicx}
\usepackage{booktabs}
\usepackage{adjustbox}
\usepackage{multirow}
\usepackage{enumitem}
\usepackage{colortbl}
\usepackage{xcolor}
\usepackage{amssymb}
\usepackage{subcaption}
\usepackage{xspace}
\usepackage{comment}
\graphicspath{{figures/}}
\definecolor{cvprblue}{rgb}{0.21,0.49,0.74}

\usepackage[
  colorlinks=true,
  citecolor=cvprblue,
  urlcolor=cvprblue
]{hyperref}

\def\tabref#1{Table~\ref{#1}}
\def\Tabref#1{Table~\ref{#1}}
\def\appref#1{Appendix~\ref{#1}}

\def\Figref#1{Figure~\ref{#1}}

\newcommand{\datasetC}{\textsc{ClimbingVideos}\xspace}
\newcommand{\wall}{\textsc{ForceWall}\xspace}
\newcommand{\ours}{\textsc{PACT}\xspace}

\title{\ours: End-to-End Learning of \\ Human Pose, Contacts, and Forces from Video}

\author{Rikhat Akizhanov$^{1}$, Yangsong Zhang$^{1}$, Nikolai Kaliazin$^{1}$, Peter Wolf$^{2}$, Yoshihiko Nakamura$^{1}$, \\
\textbf{Pascal Fua$^{3}$, Fabio Pizzati$^{1}$ \& Ivan Laptev$^{1}$} \\
$^{1}$MBZUAI \quad $^{2}$ETH Z\"urich \quad $^{3}$EPFL \\
\texttt{\{rikhat.akizhanov,ivan.laptev\}@mbzuai.ac.ae} \\
Project page: \url{https://rihat99.github.io/PACT/}
}

\confpreprint %
\begin{document}

\maketitle

\begin{figure}[h!]
\vspace{-.8cm}
\centering\includegraphics[width=0.9\linewidth]{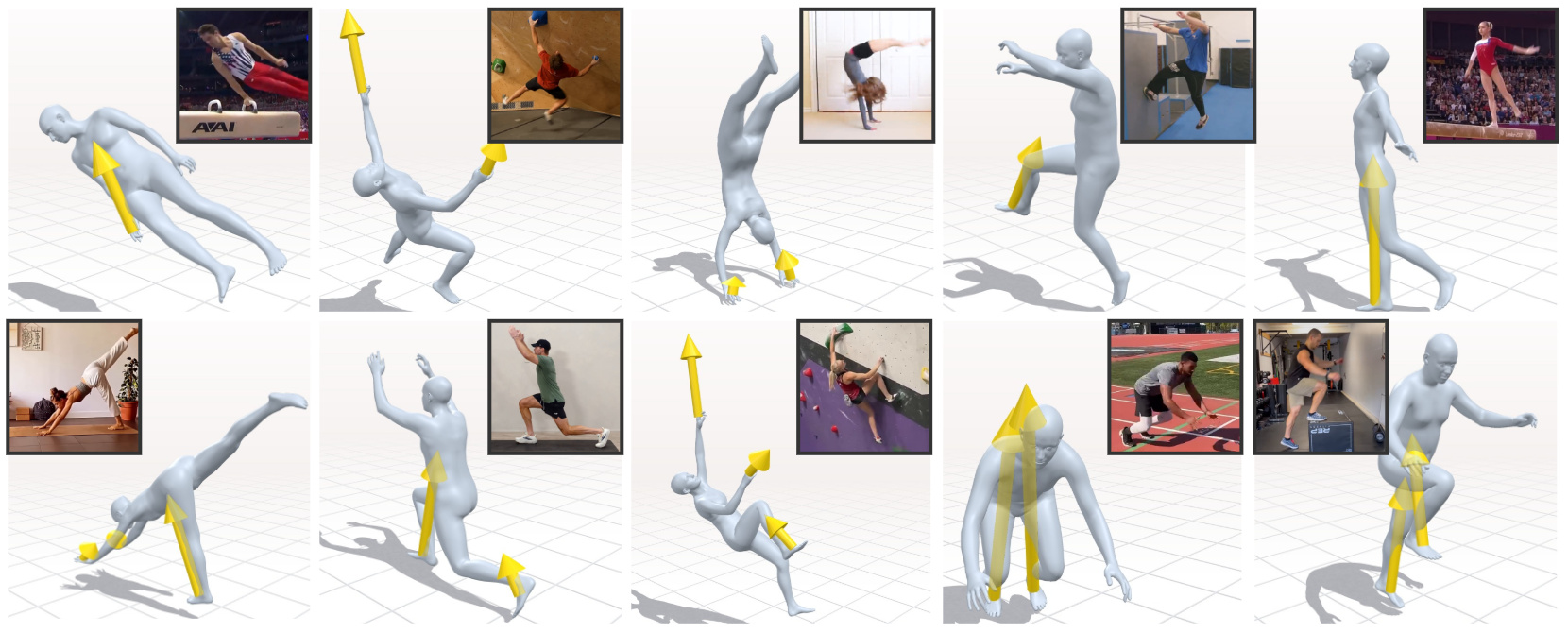}
\vspace{-.2cm}
\caption{
End-to-end joint estimation of human pose, contacts, and forces with PACT. Predictions from monocular videos show generalization across diverse activities and support configurations. White meshes show estimated poses; yellow arrows show predicted forces at estimated contacts.
\vspace{-.1cm}
}
\label{fig:teaser}
\end{figure}

\begin{abstract}
Human motion, environmental contacts, and interaction forces are governed by common physical laws, yet existing approaches typically separate visual pose reconstruction from contact and force estimation. This separation limits joint reasoning and can propagate errors between stages. We introduce PACT, an end-to-end model that jointly learns to estimate human pose, contacts and contact forces from monocular video. Our approach augments a human reconstruction foundation model with learnable contact-force tokens and a temporal transformer that integrates visual features with world-space motion. Joint prediction heads refine human poses and estimate contacts and forces, while physics-based supervision encourages consistency between the reconstructed motion and interaction forces. To address the scarcity of force annotations, we develop a data annotation pipeline that combines contact labeling with physics-based motion and force optimization, producing training supervision from synthetic and real-world videos. We also introduce a real-world climbing benchmark \wall with climbing videos and corresponding ground-truth contact forces obtained from the force sensors. Experiments demonstrate state-of-the-art contact and force estimation, outperforming staged reconstruction approaches and generalizing to interactions beyond the training distribution. These results support end-to-end joint learning as an effective approach to recovering human motion and physical interactions from video.

\end{abstract}

\section{Introduction}
\label{sec:introduction}

Recovering human movement from video requires understanding both how the body moves and how it interacts with its environment. Recent advances in monocular human reconstruction have enabled increasingly accurate estimates of body shape and pose across diverse visual conditions. However, pose alone provides an incomplete description when interactions are involved: similar body configurations can correspond to different contact states and substantially different forces. Recovering these quantities is essential for analyzing physical effort, assessing athletic performance, and learning robot behaviors from human demonstrations. This motivates reconstructing human pose, environmental contacts, and interaction forces together from monocular videos.

Human motion, contacts and interaction forces are governed by common physical laws. Motion determines the required net forces and moments, while contacts determine where external forces can act. Conversely, contact and force constraints inform plausible body trajectories. Yet existing methods typically separate visual reconstruction from interaction estimation. Optimization-based approaches recover motion and forces using independently detected contacts and initial poses~\citep{li2019_forces}. Learned dynamics models, including D\&D~\citep{dandd} and PhysPT~\citep{physpt}, operate on reconstructed motion to infer physical quantities and refine trajectories. Their dynamics modules therefore receive a restricted representation of the video: intermediate errors can propagate, while visual evidence relevant to contacts and forces may be lost.

We investigate an end-to-end formulation that jointly learns pose, contact, and force estimation, retaining visual evidence while coupling motion and forces through physics. This is particularly relevant to sports, such as climbing, where support and load shift between hands and feet. Motion alone may not uniquely determine this force distribution, whereas video provides additional evidence about body configuration and surrounding supports. Combining visual observations with world-space motion enables the model to exploit these complementary cues.

Two challenges arise. First, the model must integrate visual and temporal information: contacts depend on the body's relationship with its surroundings, while forces depend on motion over time. Human and camera motion must also be distinguished to reason about dynamics in a consistent frame. Second, force supervision is scarce. Force measurements require specialized hardware, and datasets with synchronized video, motion, contacts, and forces \mbox{cover limited activities and environments.}

In this work we introduce PACT, an end-to-end model for jointly estimating human \textbf{P}ose, contact states, \textbf{A}nd \textbf{C}on\textbf{T}act forces from monocular video. PACT builds on SAM 3D Body~\citep{sam3d_body}, retaining its pretrained visual backbone while learning modules for physical interaction estimation. Learnable contact-force tokens extract interaction-relevant visual features, and a temporal transformer combines these features with pose representations and world-space kinematics. Joint prediction heads refine the body trajectory and estimate contact probabilities and forces. A physics-based training objective couples the refined motion and predicted forces through the equations of motion, while contact predictions determine where the estimated forces are applied. This formulation preserves access to visual evidence while learning the three outputs within a shared model.

To train PACT, we develop an annotation pipeline that combines contact labeling with physics-based optimization of motion and forces. The pipeline produces supervision from synthetic sequences and real-world videos, expanding training coverage beyond datasets with directly measured forces. For evaluation, we introduce \wall, an instrumented climbing benchmark with measured contact forces and changing hand and foot supports. Experiments on \wall and established benchmarks demonstrate excellent results outperforming staged approaches. Notably our method demonstrates generalization to scenarios outside the training distribution. 
Figure~\ref{fig:teaser} shows qualitative results of \ours across diverse scenes and motions.
In summary, we propose the following contributions:\looseness=-1\smallskip\\
    \textbf{1.~End-to-end joint estimation.} We introduce PACT, a model that jointly learns human pose, contact, and force estimation from monocular video by combining pretrained visual representations, temporal reasoning, and physics-based supervision.
    \smallskip\\
    \textbf{2.~Training supervision and measured evaluation.} We develop a physics-based annotation pipeline for synthetic and real-world videos and introduce \wall, a benchmark with directly measured contact forces for evaluating interactions with changing hand and foot supports.\looseness=-1
    \smallskip\\
    \textbf{3.~Comprehensive empirical validation.} We demonstrate state-of-the-art contact and force estimation across multiple benchmarks, generalization beyond the training distribution, and ablations that validate PACT's architectural design choices and physical supervision.

\section{Related Work}
\label{sec:related}

\noindent\textbf{Human mesh recovery and world-grounded motion.}
The strongest regressors of parametric bodies~\citep{loper2015smpl,pavlakos2019smplx,ferguson2025mhr} in the camera frame~\citep{hmr2,patel2025camerahmr,wang2025prompthmr,sam3d_body} owe their accuracy to in-the-wild images whose labels were produced largely by fitting and by the models themselves~\citep{kolotouros2019spin,joo2021eft,moon2022neuralannot,patel2025camerahmr}, a scale that manual annotation cannot reach.
World-frame motion needs the camera as well, from SLAM~\citep{ye2023slahmr}, learned motion priors~\citep{shin2024wham,wang2024tram} or gravity-view coordinates~\citep{shen2024gvhmr}.
These priors are learned from motion capture, mostly AMASS~\citep{mahmood2019amass}, whose people walk, run and sit on flat ground, and most of them rely on a floor for the scale of the camera or the contacts of the feet.
We take this self-labeling route to the world frame, to contacts and to forces, with a world frame from a geometry model~\citep{wang2025vggt,wang2026vggtomega} anchored to the size of the body rather than to a floor.

\noindent\textbf{Contact estimation.}
Human-scene contact is estimated either as a dense label on the body surface, from images with BSTRO~\citep{huang2022rich}, DECO~\citep{tripathi2023deco} and InteractVLM~\citep{dwivedi2025interactvlm} or as a scene-constrained pose prior~\citep{hassan2019prox}, or as a sparse per-joint signal, typically foot contact, inside motion recovery~\citep{rempe2020contact,shin2024wham,shen2024gvhmr,wang2025prompthmr}.
The first family learns from manual or scan-derived contact annotations and the second from motion capture, both scarce and the latter dominated by feet on a floor.
Differently, we generate the labels from geometry and physics, with a scalable video-oriented pipeline, which we use to learn contact estimation end-to-end.

\noindent\textbf{Physics-based force estimation.}
\citet{brubaker2009contact} inferred contact forces and joint torques from motion, and \citet{li2019_forces,li2022forces} recover motion and contact forces by trajectory optimization given a ground plane and contacts in 2D, and \citet{gartner2022trajectory} through a physics simulator.
Physics has since served to refine poses~\citep{rempe2020contact,shimada2020physcap,xie2021physics,tripathi2023ipman}, as weak supervision for learned forces~\citep{zell2020weakly,ehsani2020force}, and as a regression target: D\&D~\citep{dandd}, PhysPT~\citep{physpt}, MusclePose~\citep{zhu2025musclepose} and ImDy~\citep{liu2025imdy} predict forces, torques or muscle activations from a reconstructed motion, trained on motion capture or simulation.
Biomechanics pipelines estimate ground reaction forces with a musculoskeletal model, by simulation from multi-view~\citep{uhlrich2023opencap} or, with learned kinematics, monocular video~\citep{gilon2026opencapmono}. GaitDynamics~\citep{tan2026gaitdynamics} assumes motion-capture poses and the ground-truth mass.
Most of these take a flat floor as the support and the feet as the contacts, \citet{li2019_forces} and \citet{gartner2022trajectory} also handle hand-held objects and scene geometry given the contacts, and all of them read the forces from a motion rather than from the image. We instead jointly predict contacts and forces from the image.

\section{Method}
\label{sec:method}

We aim to estimate human motion, contacts, and contact forces from a
monocular video. While individual frames provide evidence about body
configuration and possible contacts, forces also depend on how the
body moves over time. Hence, we use a temporal transformer to combine world-informed image features, extracted with a pretrained body reconstruction foundation model, over time. This preserves the generalization capabilities of the foundation model, learning new time-dependent outputs too (\secref{sec:architecture}). Since physics-related data annotation required for training is challenging, we use a physics-based data annotation pipeline, compatible with in-the-wild videos, in \secref{sec:data_training}.\looseness=-1

\subsection{Architecture}
\label{sec:architecture}

\subsubsection{Feature extraction}
\label{sec:feature_extraction}

\begin{figure}[t]
    \centering
    \includegraphics[width=\linewidth]{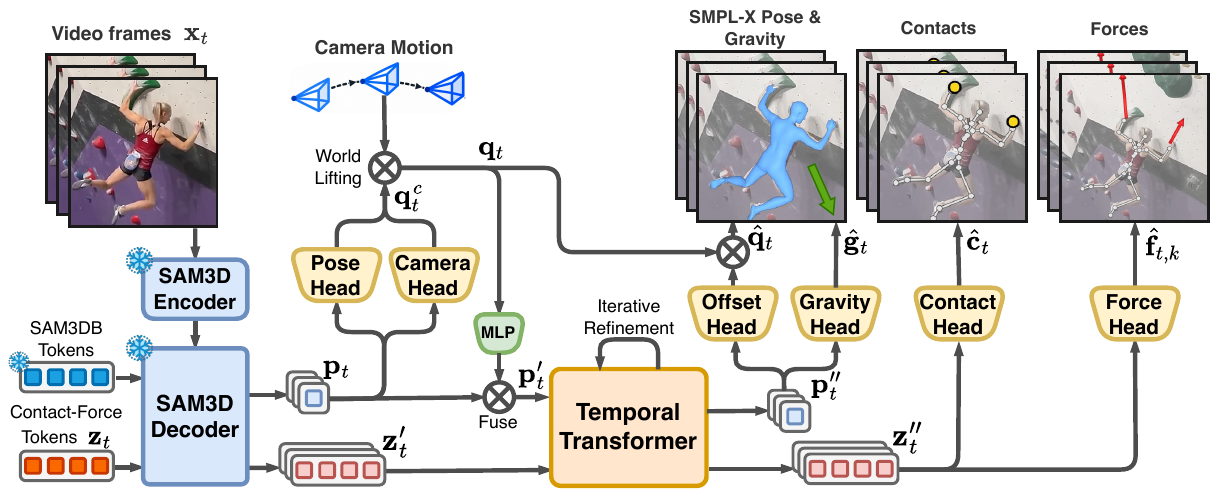}
\caption{\ours architecture. We add contact-force tokens to the frozen SAM 3D Body to account for additional outputs, that we process with a temporal transformer, alongside pose tokens enriched with world-related information. We use dedicated heads for contacts and forces, and predict pose offsets to smooth human motion learning from videos.}
\label{fig:architecture}
\end{figure}

\paragraph{Preliminaries.}
Our architecture is in \figref{fig:architecture}. We build on SAM 3D Body~\citep{sam3d_body}, which produces pose tokens that can be decoded into body
and camera parameters from arbitrary frames. These features provide an informative representation
 of human geometry, allowing us to learn generalizable contact and force
estimation without learning body reconstruction from scratch.

Let $\mathbf{x}=(\mathbf{x}_t)_{t=1}^{T}$ denote an input video. We obtain
masks $\mathbf{s}_t$ of the tracked person using SAM 3~\citep{sam3} and
process each frame independently, $\mathbf{p}_t=\operatorname{SAM3D}(\mathbf{x}_t,\mathbf{s}_t)$.
Here, $\mathbf{p}_t$ contains the pose tokens for frame $t$. We use body and
camera heads, $\mathcal{H}_{\mathrm{SMPL\text{-}X}}$ and
$\mathcal{H}_{\mathrm{camera}}$, to decode these tokens into parameters that
jointly determine the camera-frame body configuration $\mathbf{q}_t^c$.
While SAM 3D Body provides pretrained heads, we retrain these heads for compatibility with SMPL-X to reduce
engineering complexity, while keeping the pretrained backbone frozen. $\mathbf{q}_t^c$ contains root translation, root and joint
rotations, and shape, obtained from the two heads as follows:
\begin{equation}
    \begin{aligned}
        (\boldsymbol{\theta}_t^c,\boldsymbol{\beta}_t)
        =
        \mathcal{H}_{\mathrm{SMPL\text{-}X}}(\mathbf{p}_t), \quad
        \boldsymbol{\kappa}_t
        =
        \mathcal{H}_{\mathrm{camera}}(\mathbf{p}_t), \quad
        \mathbf{q}_t^c
        =
        \left(
            \mathbf{r}_t^c,
            \boldsymbol{\theta}_t^c,
            \boldsymbol{\beta}
        \right).
    \end{aligned}
    \label{eq:body_configuration}
\end{equation}
Here, $\boldsymbol{\kappa}_t$ contains the camera parameters used to
recover the camera-frame root translation $\mathbf{r}_t^c$ from the crop
box and focal length. The parameters $\boldsymbol{\theta}_t^c$ contain
the root orientation in camera coordinates and the local joint
rotations. We fix shape $\boldsymbol{\beta}$ across the video by averaging
the framewise predictions $\boldsymbol{\beta}_t$, since changes in shape
should not explain changes in motion.

\paragraph{Contact-force tokens.}
Contacts and forces depend on the human configuration, so a foundational 3D human reconstruction model should provide generalizable features for the task. Hence, we follow the principle of Visual Prompt
Tuning~\citep{visual_prompt_tuning} to introduce a learnable input embedding and corresponding output tokens on top of SAM 3D Body, retaining
the backbone weights frozen. Doing so, we benefit from generalization capabilities of the backbone. The output becomes:
\begin{equation}
    (\mathbf{p}_t,\mathbf{z}_t')
    =
    \operatorname{SAM3D}
    (\mathbf{x}_t,\mathbf{s}_t;\mathbf{z}_t).
    \label{eq:contact_force_tokens}
\end{equation}
Above, $\mathbf{z}_t$ contains learned input embeddings, while $\mathbf{z}_t'$ contains their image-dependent
outputs. Each token corresponds to one contact point: we use 35 contacts at different surface points of the human body. The points are listed in \appref{app:contacts} and shown in \figref{fig:contactpoints}.

\paragraph{World-informed poses.}
The pose tokens $\mathbf{p}_t$ and corresponding body predictions $\mathbf{q}_t^c$
are camera-relative. As such, they mix human and camera motion, having different implications for force estimation. We therefore combine these visual features with explicit
world-space kinematics to describe body motion independently of the
camera. Using camera poses from VGGT-$\Omega$~\citep{wang2026vggtomega} and video-level
scale alignment with the metric body estimate, we transform
$\mathbf{q}_t^c$ into world-space configurations $\mathbf{q}_t$. From the resulting trajectory
$\mathbf{q}=(\mathbf{q}_t)_{t=1}^{T}$, we extract SMPL-X joint
positions $\mathbf{j}_t$ relative to the root and their velocities
$\dot{\mathbf{j}}_t$, in the body frame, computed from previous and following frames using
elapsed time. Two MLPs map the pose tokens and these kinematics before
concatenation:
\begin{equation}
    \mathbf{p}_t'
    =
    \left(
        \operatorname{MLP}_{\mathrm{pose}}(\mathbf{p}_t),
        \operatorname{MLP}_{\mathrm{world}}
        (\mathbf{j}_t,\dot{\mathbf{j}}_t)
    \right).
    \label{eq:world_informed_features}
\end{equation}
Parentheses denote
concatenation. The resulting $\mathbf{p}_t'$ provides pose context while adding
world-space motion information for contact and force estimation.
\subsubsection{Predicting motion, contacts, and forces}
\label{sec:prediction}

\paragraph{Temporal aggregation.}
Similar body configurations can imply
different contact forces, making an instantaneous pose insufficient for
force estimation. We therefore aggregate the pose and contact-force
features \textit{over time}, allowing the model to relate changes in body motion
to the evolution of contacts. We introduce a temporal transformer $\mathcal{T}$ with three standard blocks. The full architecture is described in \appref{app:temporal}. We then have:
\begin{equation}
    (\mathbf{p}_t'',\mathbf{z}_t'')_{t=1}^{T}
    =
    \mathcal{T}
    \left(
        (\mathbf{p}_t',\mathbf{z}_t')_{t=1}^{T}
    \right).
    \label{eq:temporal_aggregation}
\end{equation}
 Temporal attention connects observations of the same body
region, while attention across slots allows its prediction to depend on
the whole-body configuration and the other contacts. We use RoPE embeddings~\citep{su2024rope} in temporal attention to account for
the relative temporal position of observations. The time-informed outputs remain associated with individual frames: $\mathbf{p}_t''$
describes the body at time $t$, and $\mathbf{z}_t''$ contains its
contact-force tokens.

\paragraph{Prediction heads.}
We finally map the tokens to outputs. First, we introduce two separate prediction heads $\mathcal{H}_\text{contacts}$ for contacts $\hat{\mathbf{c}}$ and $\mathcal{H}_\text{force}$ for forces $\tilde{\mathbf{f}}$, both processing the $\mathbf{z}_t''$ tokens.
\begin{equation}
    \begin{aligned}
        \hat{\mathbf{c}}_t
        =
        \sigma\!\left(
            \mathcal{H}_{\mathrm{contacts}}(\mathbf{z}_t'')
        \right), \quad
        \tilde{\mathbf{f}}_{t,k}
        =\mathcal{H}_{\mathrm{force}}(\mathbf{z}_{t,k}''), \quad
        \hat{\mathbf{f}}_{t,k}
        =\hat{c}_{t,k}\, m g\, \mathbf{R}_t\, \tilde{\mathbf{f}}_{t,k}.
    \end{aligned}
    \label{eq:contact_force_prediction}
\end{equation}
Here, $\sigma$ is a sigmoid for extracting contact probabilities. Considering that forces can exist only in presence of a contact, we use the contact probabilities $\hat{c}_{t,k}$ to gate
the forces for each contact point. The head regresses $\tilde{\mathbf{f}}_{t,k}$ in the body frame in body weights, and the root orientation $\mathbf{R}_t$ and mass $m$ convert it to world newtons (\appref{app:heads}). Gravity is a unit direction in world coordinates, with $g = 9.81$\,m/s$^2$ fixed.\looseness=-1

We also use a pose refinement head $\mathcal{H}_{\mathrm{offset}}$
used to correct SAM 3D Body pose output. Framewise prediction can indeed introduce inconsistencies affecting force estimation, which we correct with temporal information. We predict offsets in the body frame with a head $\mathcal{H}_{\mathrm{offset}}$. A gravity head $\mathcal{H}_\text{gravity}$ processes $\mathbf{p}_t''$ to extract the gravity direction $\hat{\mathbf{g}}$, taking into account tilted cameras:
\begin{equation}
\hat{\mathbf{q}}_t
=
\mathbf{q}_t
\oplus
\mathcal{H}_{\mathrm{offset}}(\mathbf{p}_t''),\quad \hat{\mathbf{g}}_{t}= \mathcal{H}_\text{gravity}(\mathbf{p}_{t}'').
\label{eq:pose_correction}
\end{equation}
The per-frame estimates $\hat{\mathbf{g}}_t$ are averaged over the clip into $\hat{\mathbf{g}}$ (\appref{app:heads}). The operator $\oplus$ adds a root translation offset and composes rotation
corrections with the root and body-joint rotations, leaving the shape
unchanged. Predicting corrections allows the model to retain the initial
reconstruction while adjusting it using neighboring observations.

\paragraph{Iterative refinement.}
Inspired by iterative refinement in SAM 3D Body~\citep{sam3d_body}, we apply the shared offset head after each transformer block, starting from $\hat{\mathbf{q}}_t^{(0)}=\mathbf{q}_t$.
At block $i$, we compute $\hat{\mathbf{q}}_t^{(i)}=\hat{\mathbf{q}}_t^{(i-1)}\oplus\mathcal{H}_{\mathrm{offset}}(\mathbf{p}_t^{\prime\prime(i)})$,
where $\mathbf{p}_t^{\prime\prime(i)}$ denotes the block's pose features.
From the refined trajectory, we recompute joint positions $\mathbf{j}_t^{(i)}$ and velocities $\dot{\mathbf{j}}_t^{(i)}$,
and add them, with the current contact and gravity estimates, to the pose features (\eqref{eq:feedback}).
These features inform the next block with intermediate estimates of refined body motion.
After the three blocks, $\hat{\mathbf{q}}_t=\hat{\mathbf{q}}_t^{(3)}$ and the offset of \eqref{eq:pose_correction} is their composition.\subsection{Data and training}
\label{sec:data_training}

\subsubsection{Data construction}
\label{sec:training_data}

\paragraph{Physics modeling.}
No dataset of in-the-wild videos carries poses, contacts and forces together, so we design a force and contacts optimization pipeline used to annotate existing videos. We consider the body as the floating-base articulated rigid body of SMPL-X in the configuration $\mathbf{q}_t$ of \secref{sec:architecture}, with segment masses and inertias derived from the shaped mesh as in PhysPT~\citep{physpt}, and $K = 35$ contact points on its surface, each with an outward normal (\appref{app:contacts}).
For a trajectory $\mathbf{q}$ and related velocity and acceleration $\dot{\mathbf{q}}, \ddot{\mathbf{q}}$,  the contact forces $\mathbf{f}_t = (\mathbf{f}_{t,k})_{k=1}^{K}$ respect the recursive Newton--Euler algorithm (RNEA)~\citep{featherstone2008rigid}:
\begin{equation}
    \boldsymbol{\tau}_t
    =
    \mathbf{M}(\mathbf{q}_t)\,\ddot{\mathbf{q}}_t
    + \mathbf{b}(\mathbf{q}_t,\dot{\mathbf{q}}_t)
    + \mathbf{G}(\mathbf{q}_t)
    - \mathbf{J}(\mathbf{q}_t)^{\top}\mathbf{f}_t,
    \label{eq:rnea}
\end{equation}
with $\mathbf{M}$, $\mathbf{b}$ and $\mathbf{G}$ the inertia, Coriolis and gravity terms and $\mathbf{J}$ the Jacobian of the contact points.
The variable $\boldsymbol{\tau}_t$ is composed of the force $\boldsymbol{\tau}^{f}_t$, the moment $\boldsymbol{\tau}^{m}_t$ at the unactuated root (vanishing when the forces explain the motion), and the joint torques $\tau_{t,j}$.

\paragraph{Pseudo labeling.}
We obtain pseudo labels for contacts and forces by assuming knowledge of the 3D scene. This is trivial in simulation. For in-the-wild videos, we track the person with SAM~3~\citep{sam3}, reconstruct the scene cloud with normals with VGGT-$\Omega$~\citep{wang2026vggtomega}, and extract 2D keypoints with Sapiens2~\citep{khirodkar2026sapiens2} and SAM 3D Body per frame. Sapiens2, specialized on keypoints, adds 2D reprojection supervision to the initial SAM 3D Body pose. With the 3D scene, we label a point in contact, $c^*_{t,k} = 1$, when it is near the scene and still, or when the root wrench of the motion cannot be explained without it. The full pipeline is in \appref{app:autolabel}.

For labeling forces, we use a kinematic objective $E_{\text{kin}}(\mathbf{q})$ derived from~\citet{li2019_forces}, with additional constraints for trajectory smoothing, depth and penetration, detailed in \appref{app:ekin}. We use it to solve for the forces and the motion together:
\begin{equation}
    \mathbf{q}^*,\mathbf{f}^*=\arg\min_{\mathbf{q},\,\mathbf{f}}\; E_{\text{kin}}(\mathbf{q}) + \sum_t \Big(
        w_F\,\frac{\|\boldsymbol{\tau}^{f}_t\|^2}{(mg)^2}
        + w_T\,\frac{\|\boldsymbol{\tau}^{m}_t\|^2}{(mg\ell)^2}
        + w_J \sum_j \mu_j\,\frac{\tau_{t,j}^2}{(mg\ell)^2}
    \Big) + E_{\text{reg}}(\mathbf{q}, \mathbf{f}),
    \label{eq:edyn}
\end{equation}
with $m$ the body mass derived from shape \citep{physpt}, $\ell = 1$\,m a unit lever that only makes the moment terms dimensionless like the force terms, $w_F$, $w_T$, $w_J$ the term weights, $\mu_j$ per-joint multipliers (\appref{app:dynamics}) and $E_{\text{reg}}$ a regularizer on the magnitude and smoothness of forces and torques. Each $\mathbf{f}_{t,k}$ is restricted to the labeled contacts and to a friction cone about the normal of its point (\appref{app:dynamics}).
By optimizing $\mathbf{q}^*$ as well, we simultaneously improve forces and obtain signal for a smoothed pose prediction. Contacts are re-read on the smoothed trajectory before the force solve (\appref{app:contacts}). Every label carries a confidence, for contacts from the stillness of the point and for forces from how well they close the equations of motion (\appref{app:confidences}).

\subsubsection{Losses}
\label{sec:losses}

\paragraph{Pose, contacts, and forces.}
The per-frame SMPL-X ($\mathcal{H}_\text{SMPL-X}$) and camera ($\mathcal{H}_\text{camera}$) heads are trained as in HMR 2.0~\citep{hmr2} with the full-frame projection of CLIFF~\citep{li2022cliff}. The gravity head $\mathcal{H}_\text{gravity}$ is supervised with the gravity of the labels, which GeoCalib~\citep{veicht2024geocalib} estimates from the video (\appref{app:gravity}).
The corrected trajectory $\hat{\mathbf{q}}_t$ of \eqref{eq:pose_correction} is supervised on its joints, rotations and world root, and on the velocities of its joints and root orientation, so that the model matches how the labels move rather than only where they are (\appref{app:losses}). With $c^*_{t,k} \in \{0,1\}$ and $\mathbf{f}^*_{t,k}$ the contact and force labels of point $k$ from \eqref{eq:edyn}, the latter rotated to the body frame and divided by $mg$, $\bar{c}^*_{t,k}$ the contacts re-read for the force solve, $w^{\text{c}}_{t,k}$ and $w^{\text{f}}_t$ their confidences, $\hat{c}_{t,k}$ the contact probability of \eqref{eq:contact_force_prediction} and $\tilde{\mathbf{f}}_{t,k} = \mathcal{H}_{\mathrm{force}}(\mathbf{z}''_{t,k})$ the force before gating, the losses are:
\begin{equation}
    \mathcal{L}_{\text{c}} = \sum_{t,k} w^{\text{c}}_{t,k}\,\operatorname{BCE}(\hat{c}_{t,k}, c^*_{t,k}),
    \qquad
    \mathcal{L}_{\text{f}} = \sum_{t,k} w^{\text{f}}_t \Big[
        \bar{c}^*_{t,k}\,\rho(\tilde{\mathbf{f}}_{t,k} - \mathbf{f}^*_{t,k})
        + (1 - \bar{c}^*_{t,k})\,\|\tilde{\mathbf{f}}_{t,k}\|
    \Big],
    \label{eq:lforce}
\end{equation}
a Huber loss $\rho$ on the points in contact and a penalty on force predicted at a free point, complemented by terms on the force magnitude and on the net force (\appref{app:losses}).

\paragraph{Physics.}
Since \ours predicts forces, contacts and poses, we can also use \eqref{eq:rnea} as a loss for physical consistency, applied to the gated world forces $\hat{\mathbf{f}}_{t,k}$ of \eqref{eq:contact_force_prediction}:
\begin{equation}
    \mathcal{L}_{\text{phys}}
    =
    \sum_t \Big(
        \lambda_F\,\rho_{\delta_F}\Big(\frac{\boldsymbol{\tau}^{f}_t}{mg}\Big)
        + \lambda_T\,\rho_{\delta_T}\Big(\frac{\boldsymbol{\tau}^{m}_t}{mg\ell}\Big)
        + \lambda_J \sum_j \mu_j\,\frac{\tau_{t,j}^2}{(mg\ell)^2}
    \Big).
    \label{eq:lphys}
\end{equation}
The full derivation is in \appref{app:physics_loss}. In the equation, $\rho_\delta$ is a pseudo-Huber loss of scale $\delta$ per component, with the weights $\lambda$ and scales $\delta$ listed in \appref{app:losses}.
A stillness term $\mathcal{L}_{\text{still}}$ penalizes the world speed of a contact point on the frames where it is labeled in contact, so that the corrected body does not slide on its supports (\eqref{eq:lstill} in \appref{app:losses}). We backpropagate the loss alongside all the supervised losses in an end-to-end fashion.

\section{Experiments}
\label{sec:experiments}

\subsection{Datasets and metrics}
\label{sec:data}

\paragraph{Training data.}
We train on BEDLAM2~\citep{bedlam2}, 18{,}076 sequences and 5.5M frames of 26{,}178 people labeled as in \secref{sec:training_data} (\appref{app:bedlam2_labels}), and on \datasetC, a new dataset of YouTube indoor climbing and bouldering videos labeled with the same pipeline. We curate it by keyword search, vision-language models on the thumbnail and the video~\citep{deitke2025molmo,clark2026molmo2}, shot detection, a full-body pose gate~\citep{jocher2023ultralytics} and one manual review (\appref{app:climbingvideos}). It holds 1{,}528 scenes from 191 videos, 4.5 hours at 24 to 60\,fps, split by source video into 1{,}419 training scenes (460k frames) and 109 test scenes (31k frames) with manual contact labels for 14 joints. We train on both.

\paragraph{Evaluation benchmarks.}
We evaluate forces and contacts against measured ground truth. For forces, the parkour sequences of \citet{li2019_forces} (Parkour) are 28 clips of vaults, muscle-ups and pull-ups with motion capture and forces from plates at the feet and a bar at the hands. The OpenCap benchmark~\citep{uhlrich2023opencap} has smartphone videos of walking, squatting, standing up from a chair and dropping from a box over force plates (OC-All), of which 22 walking trials form OC-Walking for comparison with published results~\citep{gilon2026opencapmono} (\appref{app:opencap_data}).
Our new \wall test set is a bouldering wall whose four panels, one under each limb, rest on three-axis force sensors: two climbers boulder at $0^\circ$, $16^\circ$ and $32^\circ$ overhang, 68 clips and 29.6k scored frames from one of three calibrated cameras (\figref{fig:wall_setup}, \appref{app:forcewall}). Sensors are matched to limbs with the reconstructed body, and a limb is in contact when its sensor reads 50\,N, until it drops below 30\,N. We evaluate contacts on the \datasetC test set, \wall and RICH~\citep{huang2022rich}.

\paragraph{Metrics.}
Contacts are scored by precision, recall and F1 at threshold 0.5, micro-averaged over frames and limbs.
The force of a limb is the sum of its contact points, and its error is the mean norm of the difference to the sensor over all frames in newtons (MAE). We add the mean angle between estimated and measured force where both exceed 50\,N (Angle), the Pearson correlation of the magnitudes over frames and limbs (Corr), the load-share error (Share), the mean absolute difference between a limb's share of the load and the measured one, in percentage points (pp), and the physical root residual, the mean root force and moment of \eqref{eq:rnea} left by a method's own motion and forces, in body weights (bw), which measures how consistent poses and forces are.
Poses are evaluated with MPJPE, on \wall against a multi-view fit and on Parkour with the similarity-aligned MPJPE of its evaluator (\appref{app:parkour_data}). On OpenCap we follow the benchmark protocol and report the mean absolute error of each foot's ground reaction force per axis in percent of body weight (\appref{app:opencap_data}).\looseness=-1

\subsection{Contacts}
\label{sec:exp_contacts}
We compare with BSTRO~\citep{huang2022rich}, trained on RICH, DECO~\citep{tripathi2023deco}, trained on RICH and other data, and InteractVLM~\citep{dwivedi2025interactvlm}, trained on the same data, a promptable contact detector prompted with ``scene'', as its authors recommend, and with ``wall'' for climbing. We add human reconstruction methods with contact heads, GVHMR~\citep{shen2024gvhmr} and PromptHMR~\citep{wang2025prompthmr}. In \tabref{tab:contacts} we are best on \datasetC and \wall and first on RICH in precision (\textbf{0.941}) and second in F1 (\underline{0.856}). Since BSTRO (best) is trained on RICH, \textit{we achieve almost the same performance on a domain fully out-of-distribution}, which proves the generalization of \ours.\looseness=-1

\begin{table}[t]
\caption{Contacts benchmarks. We compare on \datasetC,
the sensor contacts of \wall, and the contact annotations of RICH. We are best on climbing datasets and highly competitive on RICH against methods that have seen its training set. For \ours, RICH is completely out-of-distribution. Best is \textbf{bold}, second is \underline{underlined}.}
\label{tab:contacts}
\centering
\small
\setlength{\tabcolsep}{4pt}
\resizebox{\linewidth}{!}{%
\begin{tabular}{lccc ccc ccc}
\toprule
& \multicolumn{3}{c}{\textbf{\datasetC}}
& \multicolumn{3}{c}{\textbf{\wall}}
& \multicolumn{3}{c}{\textbf{RICH}} \\
\cmidrule(lr){2-4}
\cmidrule(lr){5-7}
\cmidrule(lr){8-10}
Method
& Prec.$\uparrow$ & Rec.$\uparrow$ & F1$\uparrow$
& Prec.$\uparrow$ & Rec.$\uparrow$ & F1$\uparrow$
& Prec.$\uparrow$ & Rec.$\uparrow$ & F1$\uparrow$ \\
\midrule

BSTRO~\citep{huang2022rich}
& 0.472 & 0.518 & 0.494
& 0.840 & 0.454 & 0.590
& 0.800 & \textbf{0.937} & \textbf{0.863} \\

DECO~\citep{tripathi2023deco}
& 0.435 & 0.594 & 0.502
& 0.821 & 0.548 & 0.658
& 0.736 & 0.751 & 0.744 \\

InteractVLM~\citep{dwivedi2025interactvlm}, ``scene''
& 0.419 & 0.503 & 0.457
& 0.830 & 0.604 & 0.699
& \underline{0.833} & \underline{0.855} & 0.844 \\

InteractVLM~\citep{dwivedi2025interactvlm}, ``wall''
& \underline{0.738} & \underline{0.849} & \underline{0.789}
& 0.820 & 0.676 & 0.741
& 0.785 & 0.669 & 0.722 \\

\midrule
GVHMR~\citep{shen2024gvhmr}
& 0.559 & 0.441 & 0.493
& \underline{0.892} & 0.680 & \underline{0.772}
& 0.814 & 0.746 & 0.778 \\

PromptHMR~\citep{wang2025prompthmr}
& 0.504 & 0.517 & 0.510
& 0.852 & \underline{0.706} & \underline{0.772}
& 0.790 & 0.814 & 0.802 \\

\midrule
\ours
& \textbf{0.948} & \textbf{0.895} & \textbf{0.921}
& \textbf{0.896} & \textbf{0.980} & \textbf{0.936}
& \textbf{0.941} & 0.784 & \underline{0.856} \\

\bottomrule
\end{tabular}%
}
\end{table}

\subsection{Force estimation}
\label{sec:exp_forces}

\paragraph{Benchmark results.}
Our closest baselines are \citet{li2019_forces} and PhysPT~\citep{physpt}. We report the published results of \citet{li2019_forces} and also feed their algorithm the SAM 3D Body poses that we start from, an improved ``\citet{li2019_forces}++''. PhysPT also receives SAM 3D Body poses. A naive inverse dynamics baseline estimates contacts with InteractVLM (``wall'') and applies \eqref{eq:rnea} to extract forces, and we report the MPJPE of its smoothed SAM 3D Body poses. We significantly outperform all baselines on both \wall and Parkour, so our method extends to different types of motion and yields robust force estimation (\Tabref{tab:forces}). In particular, the gap in Pearson correlation with the measured forces (\textbf{0.725} vs 0.243 for the second best on \wall) backs the physical validity of our output.\looseness=-1

For other scenarios, we follow the OpenCap evaluation pipeline~\citep{uhlrich2023opencap} and report in \tabref{tab:opencap} the foot force errors along the vertical (Vert), anteroposterior (AP) and mediolateral (ML) axes on OC-Walking and OC-All, against published OpenCap~\citep{uhlrich2023opencap} and OpenCap-mono~\citep{gilon2026opencapmono} results and the previous baselines, with the ground-truth mass and with the mass estimated from shape. We beat all baselines except two-view OpenCap in some scenarios (\textbf{3.5} vs 3.7 in OC-All), where \ours is monocular. The estimated mass changes the results little, suggesting robustness.

\begin{figure}[!t]
    \centering
    \includegraphics[width=1.0\linewidth]{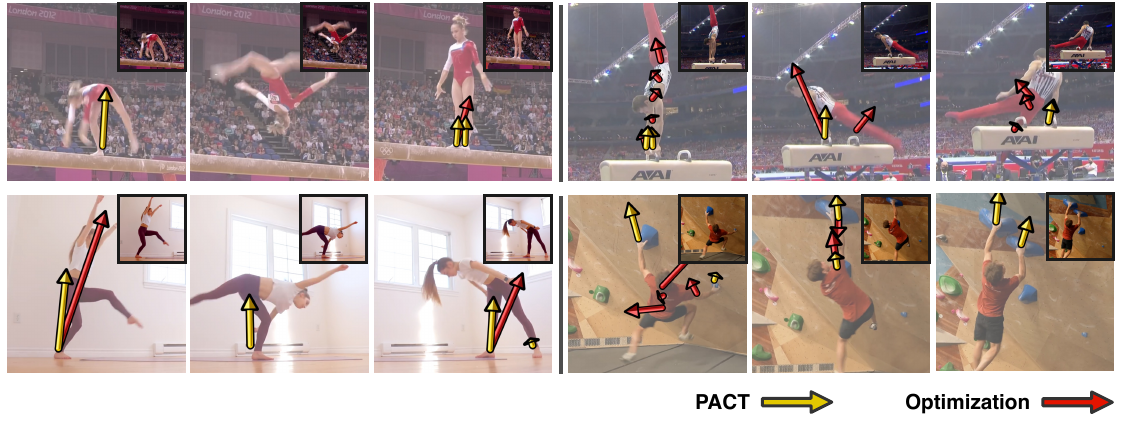}
    \caption{Qualitative comparison on in-the-wild videos. Using downloaded videos from YouTube, we generate plausible forces (yellow), while our physics-based data annotation and optimization (red) fails. This shows that \ours generalizes beyond the tool used to annotate \datasetC.}
    \label{fig:qualitatives}
\end{figure}

\paragraph{In-the-wild evaluation.}
\Figref{fig:qualitatives} shows YouTube videos of different sports. \ours yields realistic forces even without ground truth: higher forces for pushes (leftmost frames) before a jump (top) or a complex motion (bottom), and realistic contacts and forces in unusual poses holding hands (right). Our annotation pipeline (Optimization) fails completely on these videos, so by curating a dataset and training on it, \ours generalizes beyond the capabilities of its own annotation pipeline, thanks to the rich contact-force tokens benefiting from SAM 3D Body priors. More results in \appref{app:qualitative}. Please also refer to our project page for examples in motion.

\begin{table}[t]
\caption{Force estimation benchmarks. We outperform other methods in force MAE and correlation, reach the lowest moment residual and improve the poses of SAM 3D Body on Parkour.}
\label{tab:forces}
\centering
\small
\setlength{\tabcolsep}{2pt}
\resizebox{\linewidth}{!}{%
\begin{tabular}{@{}cl*{11}{c}@{}}
\toprule
& \multirow{2}{*}{\textbf{Method}}
& \multirow{2}{*}{\textbf{MPJPE [mm]}$\downarrow$}
& \multicolumn{5}{c}{\textbf{Force MAE [N]}$\downarrow$}
& \multicolumn{3}{c}{\textbf{Force agreement}}
& \multicolumn{2}{c}{\textbf{Residual}$\downarrow$} \\
\cmidrule(lr){4-8}
\cmidrule(lr){9-11}
\cmidrule(lr){12-13}
& &
& L.\ foot & R.\ foot & L.\ hand & R.\ hand & Avg.
& Angle [$^\circ$]$\downarrow$
& Corr$\uparrow$
& Share [pp]$\downarrow$
& Force [bw] & Moment [bw\,m] \\
\midrule
\multirow{5}{*}{\rotatebox[origin=c]{90}{\scriptsize{\textbf{\wall}}}}
& SAM 3D Body & \underline{42.8} & -- & -- & -- & -- & -- & -- & -- & -- & -- & -- \\
& \citet{li2019_forces}++ & 137.4 & \underline{166.2} & \underline{141.5} & \underline{128.8} & \underline{129.1} & \underline{141.4} & \underline{21.2} & \underline{0.243} & \underline{18.5} & 0.384 & 0.113 \\
& PhysPT & 86.7 & 193.8 & 147.4 & 170.3 & 159.8 & 167.8 & 21.4 & 0.169 & 28.0 & 0.785 & 0.080 \\
& Inverse dynamics & \textbf{42.7} & 200.7 & 162.2 & 166.0 & 192.4 & 180.3 & 28.8 & 0.007 & 21.9 & \textbf{0.024} & \underline{0.040} \\
& \ours & 49.1 & \textbf{98.4} & \textbf{90.4} & \textbf{72.9} & \textbf{80.4} & \textbf{85.5} & \textbf{14.4} & \textbf{0.725} & \textbf{8.7} & \underline{0.167} & \textbf{0.038} \\

\midrule

\multirow{6}{*}{\scriptsize{\rotatebox[origin=c]{90}{\textbf{Parkour}}}}
& SAM 3D Body & 68.9 & -- & -- & -- & -- & -- & -- & -- & -- & -- & -- \\
& \citet{li2019_forces} & 122.1 & 144.2 & 138.2 & 107.9 & 113.4 & 125.9 & -- & -- & -- & -- & -- \\
& \citet{li2019_forces}++ & 95.3 & \underline{132.2} & \underline{134.8} & \underline{98.7} & \underline{74.6} & \underline{110.1} & 17.1 & \underline{0.551} & \textbf{10.8} & 0.431 & \underline{0.066} \\
& PhysPT & 73.1 & 147.2 & 142.1 & 132.5 & 91.1 & 128.2 & \textbf{10.3} & 0.156 & 32.7 & 0.891 & 0.092 \\
& Inverse dynamics & \underline{68.7} & 163.2 & \textbf{126.2} & 187.1 & 205.8 & 170.6 & 23.8 & 0.527 & 18.7 & \textbf{0.042} & 0.073 \\
& \ours & \textbf{66.3} & \textbf{117.1} & 148.4 & \textbf{77.9} & \textbf{66.3} & \textbf{102.4} & \underline{16.7} & \textbf{0.712} & \underline{14.6} & \underline{0.283} & \textbf{0.044} \\

\bottomrule
\end{tabular}%
}
\end{table}

\begin{table}[t]
\caption{Evaluation on OpenCap data. We outperform all methods in all metrics except two-view OpenCap, where we are monocular. We report results with each method's estimated mass and with the measured mass, and published numbers for OpenCap and OpenCap-mono.}
\label{tab:opencap}
\centering
\small
\setlength{\tabcolsep}{4pt}
\resizebox{\linewidth}{!}{%
\begin{tabular}{@{}lll*{12}{c}@{}}
\toprule
\multirow{4}{*}{\textbf{Method}}
& \multirow{4}{*}{\textbf{Input}}
& \multirow{4}{*}{\textbf{Type}}
& \multicolumn{6}{c}{\textbf{OC-Walking}}
& \multicolumn{6}{c}{\textbf{OC-All}} \\
\cmidrule(lr){4-9}
\cmidrule(lr){10-15}
& &
& \multicolumn{3}{c}{Estimated mass}
& \multicolumn{3}{c}{True mass}
& \multicolumn{3}{c}{Estimated mass}
& \multicolumn{3}{c}{True mass} \\
\cmidrule(lr){4-6}
\cmidrule(lr){7-9}
\cmidrule(lr){10-12}
\cmidrule(lr){13-15}
& &
& Vert. & AP & ML
& Vert. & AP & ML
& Vert. & AP & ML
& Vert. & AP & ML \\
\midrule

OpenCap~\citep{uhlrich2023opencap}
& two views & simulation
& -- & -- & --
& 12.2 & \textbf{3.1} & \textbf{1.2}
& -- & -- & --
& \underline{11.4} & \textbf{3.5} & \underline{3.8} \\

OpenCap-mono~\citep{gilon2026opencapmono}
& one view & learn + optim.
& -- & -- & --
& \underline{9.7} & 4.4 & 1.7
& -- & -- & --
& -- & -- & -- \\

PhysPT~\citep{physpt}
& one view & learned
& 66.3 & \underline{8.0} & \underline{2.5}
& 64.8 & 8.0 & 2.5
& 46.0 & \underline{4.1} & \underline{3.9}
& 45.9 & 4.2 & 4.0 \\

\citet{li2019_forces}++
& one view & optimization
& \underline{33.9} & 9.0 & 4.1
& 35.0 & 8.6 & 4.0
& \underline{37.0} & 6.1 & 4.3
& 36.5 & 5.8 & 4.3 \\

\ours
& one view & learned
& \textbf{11.8} & \textbf{3.0} & \textbf{1.5}
& \textbf{7.8} & \underline{3.3} & \underline{1.6}
& \textbf{11.6} & \textbf{3.4} & \textbf{2.3}
& \textbf{9.7} & \underline{3.7} & \textbf{2.4} \\

\bottomrule
\end{tabular}%
}
\end{table}

\subsection{Ablations}
\label{sec:ablations}
\Tabref{tab:ablation} ablates first the architecture (top), trained on \datasetC only for compute reasons and tested on \wall. Adding our contributions in sequence yields the best force accuracy. The temporal transformer alone prioritizes pose estimation, but our contributions substantially increase force accuracy (\textbf{87.5} N) and Pearson correlation (\textbf{0.729}), and removing the contact head (No contact head) significantly degrades the forces, showing the benefit of multitask learning. On training data (bottom), combining BEDLAM2 with \datasetC enlarges the training distribution and compensates for suboptimal pose optimization: MPJPE improves substantially with more data (56.6 vs \textbf{49.1}) while the other metrics barely move. The temporal transformer is further ablated in \appref{app:window}.

\begin{table}[t]
\caption{Ablations on \wall.
Architecture additions are cumulative (\appref{app:ablation}),
with the physics loss completing the full model.
The training-data ablation uses the full model. The full model reaches the best force accuracy.}
\label{tab:ablation}
\centering
\footnotesize
\setlength{\tabcolsep}{4pt}
\resizebox{\linewidth}{!}{%
\begin{tabular}{l*{8}{c}}
\toprule
\multirow{2}{*}{\textbf{Setup}}
& \textbf{Pose}
& \textbf{Contacts}
& \multicolumn{4}{c}{\textbf{Forces}}
& \multicolumn{2}{c}{\textbf{Residual}} \\
\cmidrule(lr){2-2}
\cmidrule(lr){3-3}
\cmidrule(lr){4-7}
\cmidrule(lr){8-9}
& MPJPE [mm]$\downarrow$
& F1$\uparrow$
& MAE [N]$\downarrow$
& Angle [$^\circ$]$\downarrow$
& Corr$\uparrow$
& Share [pp]$\downarrow$
& Force [bw]$\downarrow$
& Moment [bw\,m]$\downarrow$ \\
\midrule
\specialrule{0.4pt}{0pt}{0pt}
\rowcolor{gray!15}
\multicolumn{9}{c}{%
\vspace{-2px}
  \small{\raisebox{-0.7px}{\textit{Architecture}}}%
  \rlap{\hspace{10em}{\scriptsize(\datasetC{} training)}}%
} \\
\midrule
Per-frame tokens and heads
& 72.9 & 0.895 & 120.1 & 22.1
& 0.472 & 12.2 & 0.471 & 0.096 \\

+ temporal transformer
& \textbf{50.5} & 0.914 & 102.2 & 17.3
& 0.649 & 10.2 & 0.422 & 0.075 \\

+ world-informed poses
& 61.2 & 0.914 & 117.2 & 21.2
& 0.504 & 11.6 & 0.291 & 0.075 \\

+ iterative refinement
& 57.7 & 0.917 & 112.0 & 19.9
& 0.578 & 11.0 & 0.286 & 0.073 \\

+ contact-force tokens
& \underline{55.4} & \textbf{0.936} & \underline{96.0} & 18.1
& \underline{0.711} & \underline{8.9} & 0.240 & 0.076 \\

+ physics loss (ours)
& 56.6 & \textbf{0.936} & \textbf{87.5} & \underline{15.0}
& \textbf{0.729} & \textbf{8.6} & \underline{0.155} & \underline{0.037} \\
\midrule
No contact head
& 60.2 & -- & 122.7 & \textbf{14.9}
& 0.562 & 10.8 & \textbf{0.130} & \textbf{0.026} \\

\midrule
\specialrule{0.4pt}{0pt}{0pt}
\rowcolor{gray!15}
\multicolumn{9}{c}{%
\vspace{-2px}
  \small{\raisebox{-0.7px}{\textit{Training data}}}%
} \\\midrule
\datasetC
& \underline{56.6} & \textbf{0.936} & \underline{87.5} & \underline{15.0}
& \textbf{0.729} & \textbf{8.6} & \textbf{0.155} & \textbf{0.037} \\

BEDLAM2
& 60.0 & 0.561 & 153.0 & 16.4
& 0.410 & 24.2 & 0.445 & 0.051 \\

Both (ours)
& \textbf{49.1} & \textbf{0.936} & \textbf{85.5} & \textbf{14.4}
& \underline{0.725} & \underline{8.7} & \underline{0.167} & \underline{0.038} \\

\bottomrule
\end{tabular}%
}
\end{table}

\section{Conclusions}
\label{sec:conclusion}
We presented \ours, an end-to-end model jointly estimating human pose, contacts and forces from monocular video. Our architecture combines pretrained visual representations with world-space motion, temporal refinement, and physics-based supervision.
We also introduced an annotation pipeline for synthetic and real-world videos, and \wall{} for evaluation against measured forces. Experiments show accurate contact and force estimation and generalization beyond the training distribution to RICH, Parkour, OpenCap and in-the-wild videos without retraining. They highlight the effectiveness of combining visual evidence and physical reasoning to recover human interactions across activities and environments.

\ifconffinal
\section*{Acknowledgements}
This work was in part supported by the Institute of Information \& communications Technology Planning \& Evaluation (IITP) grant funded by the Korean government (MSIT) (No. RS-2024-00457882, National AI Research Lab Project).
\fi

\ifconfpreprint\else
\section*{AI use statement}
Generative AI tools assisted with manuscript polishing, code implementation and debugging. Annotations are also provided by AI as described in \appref{app:autolabel}.
\fi

\section*{Ethics statement}
\datasetC is built from publicly available YouTube videos of climbers.
We release only derived annotations (tracks, body parameters, contacts, forces and confidences) together with video identifiers and time stamps, not the videos themselves, and we will honour removal requests.
The automatic gates reject clips showing children.
\wall was recorded with the informed consent of the participating climbers.
The method estimates forces from video without contact with the subject, and we do not foresee harmful uses beyond those of general human motion capture.

\section*{Reproducibility statement}
The curation rules and their thresholds are given in \secref{sec:data} and \appref{app:curation}, the labeling optimization with its objectives and weights in \secref{sec:training_data} and \appref{app:autolabel}, the model, losses and training schedule in \secref{sec:architecture}, \secref{sec:losses} and \appref{app:model}, and every metric in \appref{app:metrics}.
Code for the pipeline, the model and the evaluation scripts, together with the dataset annotations, the test-split manual labels and the benchmark recordings, will be released.

\bibliography{conference}

@String(CVPR  = {CVPR})

@String(CVPRW = {CVPRW})

@String(ICCV  = {ICCV})

@String(ECCV  = {ECCV})

@String(NeurIPS  = {NeurIPS})

@String{ICLR = {ICLR}}

@InProceedings{sam3d_body,
    author    = {Yang, Xitong and Kukreja, Devansh and Pinkus, Don and Fan, Taosha and Park, Jinhyung and Shin, Soyong and Cao, Jinkun and Liu, Jia-Wei and Ugrinovic, Nicol\'as and Sagar, Anushka and Malik, Jitendra and Feiszli, Matt and Doll\'ar, Piotr and Kitani, Kris},
    title     = {{SAM 3D Body}: Robust Full-Body Human Mesh Recovery},
    booktitle = CVPR,
    year      = {2026},
}

@misc{ferguson2025mhr,
      title={{MHR}: Momentum Human Rig},
      author={Aaron Ferguson and Ahmed A. A. Osman and Berta Bescos and Carsten Stoll and Chris Twigg and Christoph Lassner and David Otte and Eric Vignola and Fabian Prada and Federica Bogo and Igor Santesteban and Javier Romero and Jenna Zarate and Jeongseok Lee and Jinhyung Park and Jinlong Yang and John Doublestein and Kishore Venkateshan and Kris Kitani and Ladislav Kavan and Marco Dal Farra and Matthew Hu and Matthew Cioffi and Michael Fabris and Michael Ranieri and Mohammad Modarres and Petr Kadlecek and Rawal Khirodkar and Rinat Abdrashitov and Romain Prévost and Roman Rajbhandari and Ronald Mallet and Russell Pearsall and Sandy Kao and Sanjeev Kumar and Scott Parrish and Shoou-I Yu and Shunsuke Saito and Takaaki Shiratori and Te-Li Wang and Tony Tung and Yichen Xu and Yuan Dong and Yuhua Chen and Yuanlu Xu and Yuting Ye and Zhongshi Jiang},
      year={2025},
      eprint={2511.15586},
      archivePrefix={arXiv},
      primaryClass={cs.GR},
      url={https://arxiv.org/abs/2511.15586},
}

@inproceedings{sam3,
    title={{SAM} 3: Segment Anything with Concepts},
    author={Nicolas Carion and Laura Gustafson and Yuan-Ting Hu and Shoubhik Debnath and Ronghang Hu and Didac Suris Coll-Vinent and Chaitanya Ryali and Kalyan Vasudev Alwala and Haitham Khedr and Andrew Huang and Jie Lei and Tengyu Ma and Baishan Guo and Arpit Kalla and Markus Marks and Joseph Greer and Meng Wang and Peize Sun and Roman R{\"a}dle and Triantafyllos Afouras and Effrosyni Mavroudi and Katherine Xu and Tsung-Han Wu and Yu Zhou and Liliane Momeni and Rishi Hazra and Shuangrui Ding and Sagar Vaze and Francois Porcher and Feng Li and Siyuan Li and Aishwarya Kamath and Ho Kei Cheng and Piotr Dollar and Nikhila Ravi and Kate Saenko and Pengchuan Zhang and Christoph Feichtenhofer},
    booktitle=ICLR,
    year={2026},
}

@inproceedings{khirodkar2026sapiens2,
    title={{Sapiens2}},
    author={Rawal Khirodkar and He Wen and Julieta Martinez and Yuan Dong and Zhaoen Su and Shunsuke Saito},
    booktitle=ICLR,
    year={2026},
}

@inproceedings{wang2025vggt,
    author    = {Wang, Jianyuan and Chen, Minghao and Karaev, Nikita and Vedaldi, Andrea and Rupprecht, Christian and Novotny, David},
    title     = {{VGGT}: Visual Geometry Grounded Transformer},
    booktitle = CVPR,
    year      = {2025},
}

@inproceedings{wang2026vggtomega,
   author    = {Wang, Jianyuan and Chen, Minghao and Zhang, Shangzhan and Karaev, Nikita and Sch\"onberger, Johannes and Labatut, Patrick and Bojanowski, Piotr and Novotny, David and Vedaldi, Andrea and Rupprecht, Christian},
    title     = {{VGGT}-\ensuremath{\Omega}},
    booktitle = CVPR,
    year      = {2026},
}

@inproceedings{veicht2024geocalib,
  title     = {{GeoCalib}: Learning Single-image Calibration with Geometric Optimization},
  author    = {Alexander Veicht and
               Paul-Edouard Sarlin and
               Philipp Lindenberger and
               Marc Pollefeys},
  booktitle = ECCV,
  year      = {2024}
}

@inproceedings{clark2026molmo2,
    author    = {Clark, Christopher and Zhang, Jieyu and Ma, Zixian and Park, Jae Sung and Tripathi, Rohun and Lee, Sangho and Salehi, Mohammadreza and Ren, Jason and Kim, Chris Dongjoo and Yang, Yinuo and Shao, Vincent and Yang, Yue and Huang, Weikai and Gao, Ziqi and Anderson, Taira and Zhang, Jianrui and Jain, Jitesh and Stoica, George and Farhadi, Ali and Krishna, Ranjay},
    title     = {{Molmo2}: Open Weights and Data for Vision-Language Models with Video Understanding and Grounding},
    booktitle = CVPR,
    year      = {2026},
}

@inproceedings{deitke2025molmo,
    author    = {Deitke, Matt and Clark, Christopher and Lee, Sangho and Tripathi, Rohun and Yang, Yue and Park, Jae Sung and Salehi, Mohammadreza and Muennighoff, Niklas and Lo, Kyle and Soldaini, Luca and Lu, Jiasen and Anderson, Taira and Bransom, Erin and Ehsani, Kiana and Ngo, Huong and Chen, YenSung and Patel, Ajay and Yatskar, Mark and Callison-Burch, Chris and Head, Andrew and Hendrix, Rose and Bastani, Favyen and VanderBilt, Eli and Lambert, Nathan and Chou, Yvonne and Chheda, Arnavi and Sparks, Jenna and Skjonsberg, Sam and Schmitz, Michael and Sarnat, Aaron and Bischoff, Byron and Walsh, Pete and Newell, Chris and Wolters, Piper and Gupta, Tanmay and Zeng, Kuo-Hao and Borchardt, Jon and Groeneveld, Dirk and Nam, Crystal and Lebrecht, Sophie and Wittlif, Caitlin and Schoenick, Carissa and Michel, Oscar and Krishna, Ranjay and Weihs, Luca and Smith, Noah A. and Hajishirzi, Hannaneh and Girshick, Ross and Farhadi, Ali and Kembhavi, Aniruddha},
    title     = {{Molmo} and {PixMo}: Open Weights and Open Data for State-of-the-Art Vision-Language Models},
    booktitle = CVPR,
    year      = {2025},
}

@article{loper2015smpl,
      author = {Loper, Matthew and Mahmood, Naureen and Romero, Javier and Pons-Moll, Gerard and Black, Michael J.},
      title = {{SMPL}: A Skinned Multi-Person Linear Model},
      journal = {ACM Trans. Graphics (Proc. SIGGRAPH Asia)},
      month = oct,
      number = {6},
      pages = {248:1--248:16},
      publisher = {ACM},
      volume = {34},
      year = {2015}
}

@inproceedings{pavlakos2019smplx,
    author = {Pavlakos, Georgios and Choutas, Vasileios and Ghorbani, Nima and Bolkart, Timo and Osman, Ahmed A. A. and Tzionas, Dimitrios and Black, Michael J.},
    title = {Expressive Body Capture: {3D} Hands, Face, and Body From a Single Image},
    booktitle = CVPR,
    year = {2019}
}

@InProceedings{hmr2,
    author    = {Goel, Shubham and Pavlakos, Georgios and Rajasegaran, Jathushan and Kanazawa, Angjoo and Malik, Jitendra},
    title     = {{Humans in 4D}: Reconstructing and Tracking Humans with Transformers},
    booktitle = ICCV,
    month     = {October},
    year      = {2023},
}

@inproceedings{ye2023slahmr,
    title={Decoupling Human and Camera Motion from Videos in the Wild},
    author={Ye, Vickie and Pavlakos, Georgios and Malik, Jitendra and Kanazawa, Angjoo},
    booktitle=CVPR,
    month={June},
    year={2023}
}

@inproceedings{shin2024wham,
   author    = {Shin, Soyong and Kim, Juyong and Halilaj, Eni and Black, Michael J.},
    title     = {{WHAM}: Reconstructing World-grounded Humans with Accurate {3D} Motion},
    booktitle = CVPR,
    year      = {2024},
}

@inproceedings{wang2024tram,
    author = {Wang, Yufu and Wang, Ziyun and Liu, Lingjie and Daniilidis, Kostas},
    title = {{TRAM}: Global Trajectory and Motion of {3D} Humans from in-the-Wild Videos},
    year = {2024},
    booktitle = ECCV,
}

@inproceedings{shen2024gvhmr,
  title={World-Grounded Human Motion Recovery via Gravity-View Coordinates},
  author={Shen, Zehong and Pi, Huaijin and Xia, Yan and Cen, Zhi and Peng, Sida and Hu, Zechen and Bao, Hujun and Hu, Ruizhen and Zhou, Xiaowei},
  booktitle={SIGGRAPH Asia Conference Proceedings},
  year={2024}
}

@inproceedings{li2019_forces,
    author = {Li, Zongmian and Sedlar, Jiri and Carpentier, Justin and Laptev, Ivan and Mansard, Nicolas and Sivic, Josef},
    title = {Estimating {3D} Motion and Forces of Person-Object Interactions From Monocular Video},
    booktitle = CVPR,
    year = {2019}
}

@article{li2022forces,
   title={Estimating {3D} Motion and Forces of Human-Object Interactions from Internet Videos},
  author={Zongmian Li and Jiri Sedlar and Justin Carpentier and Ivan Laptev and Nicolas Mansard and Josef Sivic},
  journal={International Journal of Computer Vision},
  volume={130},
  number={2},
  pages={363--383},
  year={2022},
  publisher={Springer}
}

@inproceedings{rempe2020contact,
    author={Rempe, Davis and Guibas, Leonidas J. and Hertzmann, Aaron and Russell, Bryan and Villegas, Ruben and Yang, Jimei},
    title={Contact and Human Dynamics from Monocular Video},
    booktitle=ECCV,
    year={2020}
}

@article{shimada2020physcap,
	author = {Shimada, Soshi and Golyanik, Vladislav and Xu, Weipeng and Theobalt, Christian},
	title = {{PhysCap}: Physically Plausible Monocular {3D} Motion Capture in Real Time},
	journal = {ACM Transactions on Graphics}, 
	month = {dec},
	volume = {39},
	number = {6}, 
	articleno = {235},
	year = {2020}, 
	publisher = {ACM}, 
}

@inproceedings{xie2021physics,
    author    = {Xie, Kevin and Wang, Tingwu and Iqbal, Umar and Guo, Yunrong and Fidler, Sanja and Shkurti, Florian},
    title     = {Physics-Based Human Motion Estimation and Synthesis From Videos},
    booktitle = ICCV,
    year      = {2021},
}

@inproceedings{dandd,
    author = {Li, Jiefeng and Bian, Siyuan and Xu, Chao and Liu, Gang and Yu, Gang and Lu, Cewu},
    title = {{D\&D}: Learning Human Dynamics from Dynamic Camera},
    year = {2022},
    booktitle = ECCV,
}

@InProceedings{physpt,
    author    = {Zhang, Yufei and Kephart, Jeffrey O. and Cui, Zijun and Ji, Qiang},
    title     = {{PhysPT}: Physics-aware Pretrained Transformer for Estimating Human Dynamics from Monocular Videos},
    booktitle = CVPR,
    year      = {2024},
}

@inproceedings{zhu2025musclepose,
 author = {Zhu, Kevin and MohammadiNasrabadi, AliAsghar and Wong, Alexander and McPhee, John},
 title={{3D} Human Pose Estimation with Muscles},
 booktitle = NeurIPS,
 year = {2025},
}

@inproceedings{liu2025imdy,
    title={{ImDy}: Human Inverse Dynamics from Imitated Observations},
    author={Xinpeng Liu and Junxuan Liang and Zili Lin and Haowen Hou and Yong-Lu Li and Cewu Lu},
    booktitle=ICLR,
    year={2025},
}

@article{uhlrich2023opencap,
    author = {Uhlrich, Scott D. AND Falisse, Antoine AND Kidziński, Łukasz AND Muccini, Julie AND Ko, Michael AND Chaudhari, Akshay S. AND Hicks, Jennifer L. AND Delp, Scott L.},
    journal = {PLOS Computational Biology},
    publisher = {Public Library of Science},
    title = {{OpenCap}: Human movement dynamics from smartphone videos},
    year = {2023},
    month = {10},
    volume = {19},
    pages = {1-26},
    number = {10},
}

@article{anderson2007torque,
  title={Maximum voluntary joint torque as a function of joint angle and angular velocity: model development and application},
  author={Anderson, D. E. and Madigan, M. L. and Nussbaum, M. A.},
    journal = {Journal of Biomechanics},
    volume = {40},
    number = {14},
    pages = {3105-3113},
    year = {2007},
}

@misc{gilon2026opencapmono,
      title={{OpenCap} Monocular: {3D} Human Kinematics and Musculoskeletal Dynamics from a Single Smartphone Video}, 
      author={Selim Gilon and Emily Y. Miller and Scott D. Uhlrich},
      year={2026},
      eprint={2603.24733},
      archivePrefix={arXiv},
      primaryClass={cs.CV},
      url={https://arxiv.org/abs/2603.24733}, 
}

@inproceedings{gartner2022trajectory,
  title={Trajectory Optimization for Physics-Based Reconstruction of {3D} Human Pose from Monocular Video},
  author={G{\"a}rtner, Erik and Andriluka, Mykhaylo and Xu, Hongyi and Sminchisescu, Cristian},
  booktitle={CVPR},
  year={2022}
}

@article{tan2026gaitdynamics,
  title={{GaitDynamics}: A Generative Foundation Model for Analyzing Human Walking and Running},
  author={Tan, Tian and Van Wouwe, Tom and Werling, Keenon and Liu, C Karen and Delp, Scott and Hicks, Jennifer and Chaudhari, Akshay},
  journal={Nature Biomedical Engineering},
  year={2026},
}

@inproceedings{hassan2019prox,
author = {Hassan, Mohamed and Choutas, Vasileios and Tzionas, Dimitrios and Black, Michael J.},
title = {Resolving {3D} Human Pose Ambiguities With {3D} Scene Constraints},
booktitle = ICCV,
year = {2019}
}

@inproceedings{huang2022rich,
   author    = {Huang, Chun-Hao P. and Yi, Hongwei and H\"oschle, Markus and Safroshkin, Matvey and Alexiadis, Tsvetelina and Polikovsky, Senya and Scharstein, Daniel and Black, Michael J.},
    title     = {Capturing and Inferring Dense Full-Body Human-Scene Contact},
    booktitle = CVPR,
    year      = {2022},
}

@inproceedings{tripathi2023deco,
   author    = {Tripathi, Shashank and Chatterjee, Agniv and Passy, Jean-Claude and Yi, Hongwei and Tzionas, Dimitrios and Black, Michael J.},
    title     = {{DECO}: Dense Estimation of {3D} Human-Scene Contact In The Wild},
    booktitle = ICCV,
    year      = {2023},
}

@book{featherstone2008rigid,
  title={Rigid Body Dynamics Algorithms},
  author={Featherstone, Roy},
  year={2008},
  publisher={Springer},
  isbn={978-0-387-74314-1},
  doi={10.1007/978-1-4899-7560-7}
}

@inproceedings{kingma2015adam,
  author    = {Kingma, Diederik P. and Ba, Jimmy},
  title     = {Adam: A Method for Stochastic Optimization},
  booktitle = ICLR,
  year      = {2015},
}

@article{liu1989lbfgs,
  title   = {On the Limited Memory {BFGS} Method for Large Scale Optimization},
  author  = {Liu, Dong C. and Nocedal, Jorge},
  journal = {Mathematical Programming},
  volume  = {45},
  number  = {1},
  pages   = {503--528},
  year    = {1989}
}

@article{su2024rope,
    author = {Su, Jianlin and Ahmed, Murtadha and Lu, Yu and Pan, Shengfeng and Bo, Wen and Liu, Yunfeng},
    title = {{RoFormer}: Enhanced transformer with Rotary Position Embedding},
    year = {2024},
    address = {NLD},
    volume = {568},
    number = {C},
    journal = {Neurocomput.},
    month = feb,
    numpages = {12},
}

@misc{jocher2023ultralytics,
      title={Ultralytics {YOLO26}: Unified Real-Time End-to-End Vision Models}, 
      author={Glenn Jocher and Jing Qiu and Mengyu Liu and Shuai Lyu and Fatih Cagatay Akyon and Muhammet Esat Kalfaoglu},
      year={2026},
      eprint={2606.03748},
      archivePrefix={arXiv},
      primaryClass={cs.CV},
      url={https://arxiv.org/abs/2606.03748}, 
}

@inproceedings{dwivedi2025interactvlm,
    author    = {Dwivedi, Sai Kumar and Anti\'c, Dimitrije and Tripathi, Shashank and Taheri, Omid and Schmid, Cordelia and Black, Michael J. and Tzionas, Dimitrios},
    title     = {{InteractVLM}: {3D} Interaction Reasoning from {2D} Foundational Models},
    booktitle = CVPR,
    year      = {2025},
}

@inproceedings{wang2025prompthmr,
    author    = {Wang, Yufu and Sun, Yu and Patel, Priyanka and Daniilidis, Kostas and Black, Michael J. and Kocabas, Muhammed},
    title     = {{PromptHMR}: Promptable Human Mesh Recovery},
    booktitle = CVPR,
    year      = {2025},
}

@article{keller2023skel,
  title = {From Skin to Skeleton: Towards Biomechanically Accurate {3D} Digital Humans},
  author = {Keller, Marilyn and Werling, Keenon and Shin, Soyong and Delp, Scott and Pujades, Sergi and Liu, C. Karen and Black, Michael J.},
  journal = {ACM Transaction on Graphics},
  volume = {42},
  number = {6},
  pages = {253:1-253:15},
  month = dec,
  year = {2023},
  month_numeric = {12}
}

@inproceedings{patel2025camerahmr,
    title={{CameraHMR}: Aligning People with Perspective}, 
    author={Patel, Priyanka and Black, Michael J.}, 
    booktitle={International Conference on {3D} Vision (3DV)}, 
    year={2025} 
}

@techreport{dempster1955space,
  title={Space requirements of the seated operator},
  author={Dempster, Wilfred T},
  journal={WADC Technical report},
  volume={55},
  pages={159},
  year={1955},
  publisher={Wright-Patterson Air Force Base Ohio}
}

@inproceedings{mahmood2019amass,
  title = {{AMASS}: Archive of Motion Capture as Surface Shapes},
  author = {Mahmood, Naureen and Ghorbani, Nima and Troje, Nikolaus F. and Pons-Moll, Gerard and Black, Michael J.},
  booktitle = ICCV,
  year = {2019},
}

@inproceedings{moon2022neuralannot,
    author = {Moon, Gyeongsik and Choi, Hongsuk and Lee, Kyoung Mu},  
    title = {{NeuralAnnot}: Neural Annotator for {3D} Human Mesh Training Sets},  
    booktitle = CVPRW,  
    year = {2022}  
}

@inproceedings{joo2021eft,
  title={Exemplar Fine-Tuning for {3D} Human Pose Fitting Towards In-the-Wild {3D} Human Pose Estimation},
  author={Joo, Hanbyul and Neverova, Natalia and Vedaldi, Andrea},
  booktitle={3DV},
  year={2021}
}

@inproceedings{brubaker2009contact,
  author={Brubaker, Marcus A. and Sigal, Leonid and Fleet, David J.},
  booktitle=ICCV, 
  title={Estimating contact dynamics}, 
  year={2009},
}

@inproceedings{zell2020weakly,
author = {Zell, Petrissa and Rosenhahn, Bodo and Wandt, Bastian},
title = {Weakly-Supervised Learning of Human Dynamics},
year = {2020},
booktitle = ECCV,
}

@inproceedings{ehsani2020force,
    author = {Ehsani, Kiana and Tulsiani, Shubham and Gupta, Saurabh and Farhadi, Ali and Gupta, Abhinav},
    title = {Use the Force, {Luke}! Learning to Predict Physical Forces by Simulating Effects},
    booktitle = CVPR,
    year = {2020}
}

@inproceedings{visual_prompt_tuning,
  title={Visual Prompt Tuning},
  author={Jia, Menglin and Tang, Luming and Chen, Bor-Chun and Cardie, Claire and Belongie, Serge and Hariharan, Bharath and Lim, Ser-Nam},
  booktitle=ECCV,
  year={2022}
}

@inproceedings{tripathi2023ipman,
    title = {{3D} Human Pose Estimation via Intuitive Physics},
    author = {Tripathi, Shashank and M{\"u}ller, Lea and Huang, Chun-Hao P. and Taheri, Omid
    and Black, Michael J. and Tzionas, Dimitrios},
    booktitle = CVPR,
    year = {2023}
}

@inproceedings{li2022cliff,
    author = {Li, Zhihao and Liu, Jianzhuang and Zhang, Zhensong and Xu, Songcen and Yan, Youliang},
    title = {{CLIFF}: Carrying Location Information in Full Frames into Human Pose and Shape Estimation},
    year = {2022},
    booktitle = ECCV,
}

@inproceedings{kolotouros2019spin,
    author = {Kolotouros, Nikos and Pavlakos, Georgios and Black, Michael J. and Daniilidis, Kostas},
    title = {Learning to Reconstruct {3D} Human Pose and Shape via Model-Fitting in the Loop},
    booktitle = ICCV,
    year = {2019}
}

@inproceedings{bedlam2,
  title={{BEDLAM}2.0: Synthetic humans and cameras in motion},
  author={Joachim Tesch and Giorgio Becherini and Prerana Achar and Anastasios Yiannakidis and Muhammed Kocabas and Priyanka Patel and Michael J. Black},
  booktitle=NeurIPS,
  year={2025}
}

@misc{aharon2022bot,
      title={{BoT-SORT}: Robust Associations Multi-Pedestrian Tracking}, 
      author={Nir Aharon and Roy Orfaig and Ben-Zion Bobrovsky},
      year={2022},
      eprint={2206.14651},
      archivePrefix={arXiv},
      primaryClass={cs.CV},
      url={https://arxiv.org/abs/2206.14651}, 
}

@article{HOLZBAUR20072442,
    title = {Moment-generating capacity of upper limb muscles in healthy adults},
    journal = {Journal of Biomechanics},
    volume = {40},
    number = {11},
    pages = {2442-2449},
    year = {2007},
    issn = {0021-9290},
    author = {Katherine R.S. Holzbaur and Scott L. Delp and Garry E. Gold and Wendy M. Murray},
}

@article{DELP19961371,
    title = {Maximum isometric moments generated by the wrist muscles in flexion-extension and radial-ulnar deviation},
    journal = {Journal of Biomechanics},
    volume = {29},
    number = {10},
    pages = {1371-1375},
    year = {1996},
    issn = {0021-9290},
    author = {Scott L. Delp and Anita E. Grierson and Thomas S. Buchanan},
}
\bibliographystyle{conference}

\appendix
\section{Annotation pipeline details}
\label{app:autolabel}

\begin{figure}[t]
    \centering
    \includegraphics[width=0.85\linewidth]{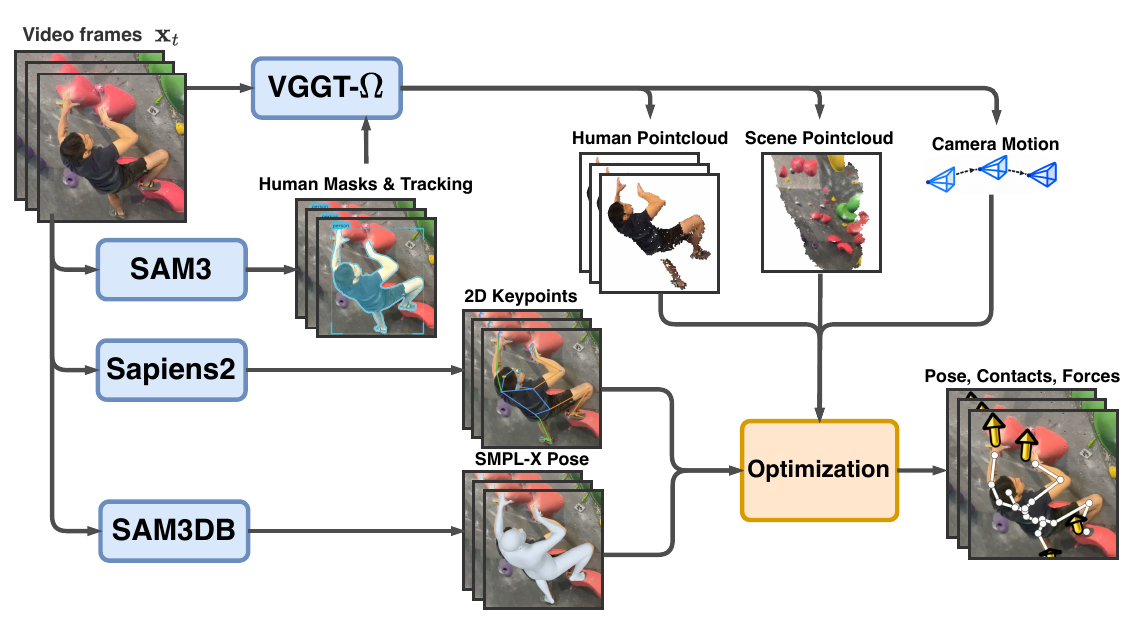}
    \caption{\textbf{Annotation pipeline.} Foundation models reconstruct the scene, the cameras and an initial body, and a physics-based optimization labels the contacts, the motion and the forces.}
    \label{fig:annotation_pipeline}
\end{figure}

This appendix details the pipeline of \secref{sec:training_data} (\figref{fig:annotation_pipeline}) and keeps the notation of \secref{sec:method}: $\mathbf{q}_t = (\mathbf{r}_t, \boldsymbol{\theta}_t, \boldsymbol{\beta})$ is the world configuration of frame $t$, $\mathbf{q}^0 = (\mathbf{q}^0_t)_{t=1}^{T}$ the initial trajectory obtained by lifting the per-frame SAM 3D Body estimates $\mathbf{q}^c_t$ to the world, $\mathbf{j}_{t,j}$ the world position of joint $j$, unlike the root-relative $\mathbf{j}_t$ of \secref{sec:method}, and $\Pi_t$ the projection of camera $t$.
For contact point $k$ we write $\mathbf{v}_{t,k}(\mathbf{q}_t)$ and $\mathbf{n}_{t,k}(\mathbf{q}_t)$ for its world position and outward normal, and $\bar{\mathbf{v}}_{t,k}$ and $\bar{\mathbf{n}}_{t,k}$ for the nearest scene point and its normal.
The operator $\Delta$ is the forward time difference, $\Delta(\cdot)_t = (\cdot)_{t+1} - (\cdot)_t$, and $\Delta t$ the frame spacing.

\subsection{Reconstruction and scale}
\label{app:reconstruction}

SAM~3~\citep{sam3} tracks the person from a text prompt and gives the masks $\mathbf{s}_t$.
VGGT-$\Omega$~\citep{wang2026vggtomega} returns camera poses, intrinsics and depth for up to 500 frames at a resolution of 512, and the depth is back-projected and split with the person mask into a static scene cloud with normals and a per-frame human cloud $\mathcal{P}_t$.
Sapiens2~\citep{khirodkar2026sapiens2} detects 308 keypoints $\mathbf{y}_{t,i}$ with scores $\omega_{t,i}$, and the per-frame SAM 3D Body~\citep{sam3d_body} estimate, converted to SMPL-X by fitting and lifted with the camera poses, gives $\mathbf{q}^0$.

\paragraph{Scale.}
The reconstruction of VGGT-$\Omega$ is up to scale, and the body serves as the ruler.
In each frame we take the median, over the front-facing vertices of the body, of the log ratio between the depth $D^{\text{body}}$ of the projected vertex and the depth $D^{\text{cloud}}$ of the human cloud at the same pixel, and pool the frames with a robust mean,
\begin{equation}
    \log \alpha = \operatorname{mean}_t \operatorname{median}_{\text{vertices}} \big( \log D^{\text{body}}_t - \log D^{\text{cloud}}_t \big),
    \label{eq:scale}
\end{equation}
then refine $\alpha$ by alternating a body fit and a scale update with the shape frozen.
The scale rescales the camera translations and both clouds, and $\mathbf{q}^0$ is expressed in the resulting world.
The shape $\boldsymbol{\beta}$ is averaged over the person's valid frames, so the height of the body comes from SAM 3D Body alone, since a single view cannot tell a large person far away from a small one nearby.

\subsection{Contact points and labeling rule}
\label{app:contacts}

\begin{figure}[t]
\begin{center}
\includegraphics[width=0.9\linewidth]{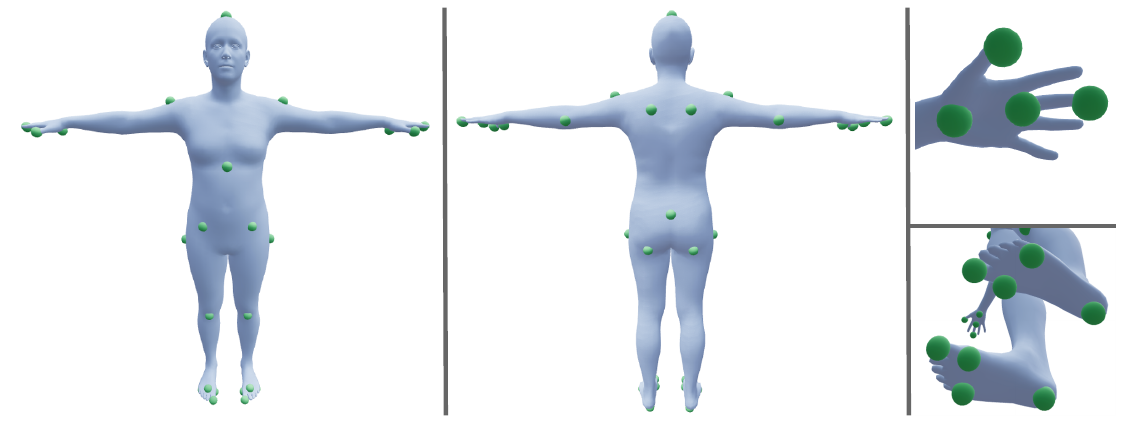}
\end{center}
\caption{\textbf{Contact points.} The $K = 35$ contact points on the SMPL-X body, front and back, with the hand and foot points enlarged. Each point carries a frame whose normal $\mathbf{n}_{t,k}$ defines its friction cone.}
\label{fig:contactpoints}
\end{figure}

The $K = 35$ points (\figref{fig:contactpoints}) cover the heels, balls and toes of the feet, the palms, fingers and thumbs of the hands, the knees, elbows, shoulders, back, hips, sit bones, sacrum, pelvis, chest and head, each attached to a template vertex with the outward normal of the mesh around it.
Point $k$ is described by its signed distance to the scene, read along the normal of the nearest scene point, and by its drift over a window of $0.35$\,s spanned by $2h$ frames,
\begin{equation}
    d_{t,k} = \big( \mathbf{v}_{t,k} - \bar{\mathbf{v}}_{t,k} \big)^{\top} \bar{\mathbf{n}}_{t,k},
    \qquad
    \xi_{t,k} = \| \mathbf{v}_{t+h,k} - \mathbf{v}_{t-h,k} \|,
    \label{eq:sdf}
\end{equation}
with the distance read on the current body and the drift on the trajectory before any contact term acts on it.

\paragraph{Geometric rule.}
A point is a geometric contact when it is near the scene, entering at $d_{t,k} < 3$\,cm and leaving at $d_{t,k} > 3.5$\,cm, and still, $\xi_{t,k} < 8$\,cm.

\paragraph{Physical test.}
The geometric rule misses a loaded limb that slides or that the scene cloud does not cover, so we also test whether the motion can be explained without a point.
Let $\boldsymbol{\tau}^{0}_t$ be the root force and moment of \eqref{eq:rnea} evaluated at $\mathbf{f}_t = 0$, that is, the wrench that the contacts must provide, and $\mathbf{J}^{\text{r}}_{t,k}$ the root block of the Jacobian of point $k$.
At frozen pose, every candidate point, near the scene or still, receives a force $\mathbf{f}_{t,k}$ in its friction cone $\mathcal{C}_{t,k}$ of \eqref{eq:cone}, and the forces of all candidates solve
\begin{equation}
    \min_{\mathbf{f}_{t,k} \in \mathcal{C}_{t,k}}\;
    \frac{1}{(mg)^2} \Big\| \boldsymbol{\tau}^{0}_t - \sum_k \mathbf{J}^{\text{r}\top}_{t,k} \mathbf{f}_{t,k} \Big\|^2
    + \sum_k \pi_{t,k}\, \frac{\| \mathbf{f}_{t,k} \|^2}{(mg)^2},
    \label{eq:witness}
\end{equation}
with the moment rows divided by a lever of $0.2$\,m, where the price $\pi_{t,k}$ of a point grows with its distance to the scene and with its speed, so that the wrench is explained by the plausible points first.
A point is a physical contact when its load enters above 2\% of body weight and stays above 1\%.

\paragraph{Label.}
The label is the union of the two rules, $c^*_{t,k} = 1$ when either fires, after which spans shorter than $0.1$\,s are removed and nearby spans of the same point are merged.
Knees are never labeled, since their geometric precision is low, and heels are labeled by geometry only, since a flat foot can route its whole load through the ball.
The labels are read on the result of each kinematic pass of \appref{app:ekin}.
The first readout is the contact label $c^*$ that supervises the model, and the second, $\bar{c}^*$, taken after the pass that pulls each labeled point towards the scene, selects the points that carry a force in the dynamics pass.

\subsection{Kinematic objective}
\label{app:ekin}

The kinematic objective of \eqref{eq:edyn} is a sum of image, prior, smoothness and scene terms,
\begin{equation}
    E_{\text{kin}}(\mathbf{q}) = E_{\text{img}}(\mathbf{q}) + E_{\text{prior}}(\mathbf{q}) + E_{\text{smooth}}(\mathbf{q}) + E_{\text{scene}}(\mathbf{q}),
    \label{eq:ekin}
\end{equation}
whose weights in each pass are listed in \tabref{tab:kinweights}.
Each sum below is averaged over its active terms, the keypoint term over the sum of the scores, and the weights refer to these averages at 30\,fps.
The variables are the root translation $\mathbf{r}_t$ and the root and joint rotations $\boldsymbol{\theta}_t$ of every frame, as unit quaternions, with the shape $\boldsymbol{\beta}$ frozen.

\paragraph{Image terms.}
With $\mathbf{o}_{t,i}(\mathbf{q}_t)$ the landmarks of the body that the detector sees, $\mathcal{V}(\mathbf{q}_t)$ its visible mesh vertices, $\operatorname{CD}$ a saturated Chamfer distance and $\rho_{\text{GM}}$ a Geman--McClure kernel that saturates on outliers,
\begin{equation}
    \begin{aligned}
    E_{\text{img}}(\mathbf{q}) = \sum_t \Big[ &
        w_{\text{init}} \sum_j \rho_{\text{GM}}\big( \Pi_t(\mathbf{j}_{t,j}) - \Pi_t(\mathbf{j}^0_{t,j}) \big)
        + w_{\text{2D}} \sum_i \omega_{t,i}\, \rho_{\text{GM}}\big( \Pi_t(\mathbf{o}_{t,i}) - \mathbf{y}_{t,i} \big) \\
        & + w_{\text{ch}}\, \operatorname{CD}\big( \mathcal{V}(\mathbf{q}_t), \mathcal{P}_t \big)
    \Big].
    \end{aligned}
    \label{eq:eimg}
\end{equation}
The Chamfer term fixes the depth of the body in the first passes.

\paragraph{Priors.}
The pose and orientation priors keep the rotations near the initial estimate through a smooth quaternion distance,
\begin{equation}
    E_{\text{prior}}(\mathbf{q}) = \sum_t \Big[
        w_{\text{pose}} \sum_{j \geq 1} \big( 1 - \langle \boldsymbol{\theta}_{t,j}, \boldsymbol{\theta}^0_{t,j} \rangle^2 \big)
        + w_{\text{orient}} \big( 1 - \langle \boldsymbol{\theta}_{t,0}, \boldsymbol{\theta}^0_{t,0} \rangle^2 \big)
    \Big],
    \label{eq:eprior}
\end{equation}
where $j = 0$ is the root, whose translation stays free.

\paragraph{Smoothness.}
The smoothness terms act on the second to fourth time differences, which removes the flicker of per-frame estimates while keeping the small movements that load and unload a limb, and a last term penalizes the velocity of each joint along its camera ray $\mathbf{u}_{t,j}$, the direction that a single view constrains least:
\begin{equation}
    E_{\text{smooth}}(\mathbf{q}) = \sum_t \Big[
        \sum_{n=2}^{4} \Big( w^{\text{r}}_n \big\| \Delta^n (\mathbf{r}_t, \boldsymbol{\theta}_{t,0}) \big\|^2
        + w^{\theta}_n \sum_{j \geq 1} \big\| \Delta^n \boldsymbol{\theta}_{t,j} \big\|^2 \Big)
        + w_{\text{rad}} \sum_j \big( \Delta \mathbf{j}_{t,j}^{\top}\, \mathbf{u}_{t,j} \big)^2
    \Big],
    \label{eq:esmooth}
\end{equation}
with the differences of the root pose taken in $\mathfrak{se}(3)$ and those of the joint rotations in $\mathfrak{so}(3)$.

\paragraph{Scene terms.}
The scene terms, a Huber penalty $\rho$ on penetration, an attraction of the labeled points to the nearest surface, an acquisition term and a stillness term, act on the contact points under the current labels $c^*_{t,k}$, provisional in the first pass and those of \appref{app:contacts} afterwards:
\begin{equation}
    \begin{aligned}
    E_{\text{scene}}(\mathbf{q}) = \sum_{t,k} \Big[ &
        w_{\text{pen}}\, \rho\big( \max(0, -d_{t,k}) \big)
        + c^*_{t,k}\, w_{\text{att}} \max(0, d_{t,k}) \\
        & + c^*_{t,k}\, w_{\text{acq}}\, \rho_{\delta}\big( \operatorname{dist}(\mathbf{v}_{t,k}, \mathcal{Q}_{t,k}) \big)
        + c^*_{t,k}\, c^*_{t+1,k}\, w_{\text{still}}\, \| \Delta \mathbf{v}_{t,k} \|^2
    \Big].
    \end{aligned}
    \label{eq:escene}
\end{equation}
Acquisition uses the pseudo-Huber kernel $\rho_{\delta}$ of \eqref{eq:pseudohuber}, applied to the scalar distance, towards a patch $\mathcal{Q}_{t,k}$ of scene points near the projection of the point, fixed during the fit, which lets the second pass reach a hold that the first pass missed.

\paragraph{Passes.}
The optimization runs in four passes with the weights of \tabref{tab:kinweights}.
The root pass fits the root alone.
The first kinematic pass fits all degrees of freedom with Adam~\citep{kingma2015adam}, with the finger rotations frozen, adds the scene terms under provisional contacts from a looser geometric rule and ends with a few iterations of L-BFGS~\citep{liu1989lbfgs}.
The second kinematic pass releases the fingers and keeps the scene terms from the start under the labels of \appref{app:contacts}, and the dynamics pass of \appref{app:dynamics} adds the forces.

\begin{table}[h]
\caption{\textbf{Auto-labeling weights.} Terms and weights of \eqref{eq:ekin} to \eqref{eq:escene} and of \eqref{eq:edyn} and \eqref{eq:ereg} in the root pass, the two kinematic passes and the dynamics pass. Grouped cells list the weights of the second, third and fourth differences for smoothness, of the first to third differences for the forces and of the first and second for the torques.}
\label{tab:kinweights}
\begin{center}
\footnotesize
\setlength{\tabcolsep}{4pt}
\resizebox{\linewidth}{!}{%
\begin{tabular}{llrrrr}
\toprule
Term & Weight & Root & Pass 1 & Pass 2 & Dynamics \\
\midrule
Reprojection to the initial joints & $w_{\text{init}}$ & 30 & 30 & 30 & 30 \\
Reprojection to the Sapiens2 keypoints & $w_{\text{2D}}$ & 8 & 8 & 8 & 8 \\
Chamfer distance to the human depth cloud & $w_{\text{ch}}$ & 10 & 10 & 0 & 0 \\
Pose prior & $w_{\text{pose}}$ & 300 & 300 & 300 & 300 \\
Root orientation prior & $w_{\text{orient}}$ & 50 & 50 & 50 & 50 \\
Smoothness of the root pose & $w^{\text{r}}_n$ & 24 / 300 / 300 & 400 / 300 / 300 & 24 / 300 / 300 & 24 / 300 / 300 \\
Smoothness of the joint rotations & $w^{\theta}_n$ & 48 / 500 / 500 & 800 / 500 / 500 & 48 / 500 / 500 & 48 / 500 / 500 \\
Radial velocity along the camera ray & $w_{\text{rad}}$ & 0 & 400 & 100 & 100 \\
Penetration & $w_{\text{pen}}$ & 0 & 12000 & 12000 & 12000 \\
Contact attraction & $w_{\text{att}}$ & 0 & 10 & 0 & 20 \\
Contact acquisition & $w_{\text{acq}}$ & 0 & 0 & 1000 & 0 \\
Contact stillness & $w_{\text{still}}$ & 0 & 500 & 4000 & 4000 \\
\midrule
Root wrench, force / moment & $w_F$ / $w_T$ & -- & -- & -- & 50 / 300 \\
Joint torques, scaled by $\mu_j$ & $w_J$ & -- & -- & -- & 30 \\
Force magnitude & $w_{\text{mag}}$ & -- & -- & -- & 1 \\
Force smoothness & $w^{\Delta f}_n$ & -- & -- & -- & 120 / 640 / 400 \\
Torque smoothness & $w^{\Delta \tau}_n$ & -- & -- & -- & 8 / 30 \\
\bottomrule
\end{tabular}}
\end{center}
\end{table}

\subsection{Force model and dynamics pass}
\label{app:dynamics}

\paragraph{Multibody.}
The body of \eqref{eq:rnea} has 52 segments, one per SMPL-X joint except the jaw and eyes, whose parts are merged into the head, whose masses and inertias are integrated from the shaped mesh as in PhysPT~\citep{physpt} with the tissue densities of \citet{dempster1955space}, and $m$ is their sum.
The velocities $\dot{\mathbf{q}}_t$ and accelerations $\ddot{\mathbf{q}}_t$ are central differences on the manifold, and gravity points along the estimated down direction of \appref{app:gravity}.
Over the corpus the estimated mass averages 64\,kg (46 to 86\,kg), and on \wall it is 62.2\,kg for the 72.5\,kg climber and 51.5\,kg for the 48\,kg climber, which we attribute to a shape estimator that compresses towards the mean of its training population.

\paragraph{Friction cone.}
Each force lives on its contact point and is restricted to a pyramidal friction cone of half-angle $45^\circ$ about the inward normal $-\mathbf{n}_{t,k}$, spanned by four edges $\mathbf{e}_{t,k,\nu}$ with non-negative coefficients,
\begin{equation}
    \mathbf{f}_{t,k} \in \bar{c}^*_{t,k}\, \mathcal{C}_{t,k},
    \qquad
    \mathcal{C}_{t,k} = \Big\{ \sum_{\nu=1}^{4} a_{\nu}\, \mathbf{e}_{t,k,\nu} : a_{\nu} \geq 0 \Big\},
    \label{eq:cone}
\end{equation}
with the edges evaluated at the pose that starts the pass and held fixed, so that the body cannot rotate to make a force cheap.
The force enters \eqref{eq:rnea} through the Jacobian of its point, as a force and a lever moment on the parent joint, so that moving the body changes the physics.

\paragraph{Regularizer.}
The term $E_{\text{reg}}$ of \eqref{eq:edyn} penalizes the magnitude of each force, its first three time differences and the first two time differences of the joint torques,
\begin{equation}
    E_{\text{reg}}(\mathbf{q}, \mathbf{f}) = \sum_{t,k} \Big[
        w_{\text{mag}} \frac{\|\mathbf{f}_{t,k}\|^2}{(mg)^2}
        + \sum_{n=1}^{3} w^{\Delta f}_n \frac{\| \Delta^n \mathbf{f}_{t,k} \|^2}{(mg)^2}
    \Big]
    + \sum_{t,j} \sum_{n=1}^{2} w^{\Delta \tau}_n \frac{(\Delta^n \tau_{t,j})^2}{(mg\ell)^2},
    \label{eq:ereg}
\end{equation}
where a force difference is taken only where the contact persists over its stencil, with half the weight on the component across the mean direction of the force, so that a force may turn more freely than it may change magnitude.

\paragraph{Torque multipliers.}
\Tabref{tab:torque} gives the multipliers $\mu_j$ of the joint-torque term, 200\,N\,m divided by an approximate maximum voluntary torque of each joint group \citep{anderson2007torque, HOLZBAUR20072442, DELP19961371}.
This term resolves the split of the load between limbs that the root wrench alone leaves undetermined: a load that the body could carry with either limb goes where the weighted torque is cheapest.
Climbers are stronger in the fingers than the population these numbers describe, so the values are coarse.

\begin{table}[h]
\caption{\textbf{Torque multipliers.} The multipliers $\mu_j$ per joint group and the maximum voluntary torque they encode.}
\label{tab:torque}
\begin{center}
\small
\setlength{\tabcolsep}{4pt}
\resizebox{\linewidth}{!}{%
\begin{tabular}{lccccccc}
\toprule
 & hip, knee, spine & ankle, foot & neck, head & collar, shoulder & elbow & wrist & fingers \\
\midrule
Max.\ torque [N\,m] & 200 & 118 & 33 & 80 & 61 & 12 & 8 \\
$\mu_j$ & 1.0 & 1.7 & 6.0 & 2.5 & 3.3 & 17 & 25 \\
\bottomrule
\end{tabular}}
\end{center}
\end{table}

\paragraph{Solver.}
The dynamics pass first solves the cone coefficients $a_{\nu}$ with L-BFGS at the frozen pose of the second kinematic pass, then pose and forces jointly with the kinematic terms of \eqref{eq:ekin} still active, which gives $\mathbf{q}^*$ and $\mathbf{f}^*$.
The labeled force $\mathbf{f}^*_{t,k}$ is stored in the world in newtons and, for training, converted to units of body weight in the body frame of the input trajectory $\mathbf{q}_t$ (\appref{app:losses}).

\subsection{Label confidences}
\label{app:confidences}

The contact confidence $w^{\text{c}}_{t,k}$ is a logistic function of the log drift $\xi_{t,k}$ of the point over the stillness window and of the log time to the nearest change of its label, with one intercept per body part and separate coefficients for contact and free labels.
The force confidence of a frame is
\begin{equation}
    w^{\text{f}}_t = \psi(\epsilon^{f}_t)\, \psi(\epsilon^{m}_t)\, \psi(\epsilon^{o}_t),
    \qquad
    \psi(x) = \frac{1}{1 + (x/x_0)^2},
    \label{eq:wforce}
\end{equation}
where $\epsilon^{f}_t = \|\boldsymbol{\tau}^{f}_t\| / (mg)$ and $\epsilon^{m}_t = \|\boldsymbol{\tau}^{m}_t\| / (mg \cdot 0.2\,\text{m})$ are the residual root force and moment of \eqref{eq:rnea} at $(\mathbf{q}^*, \mathbf{f}^*)$, $\epsilon^{o}_t$ the excess of the summed force magnitudes over three body weights, in body weights, and $x_0 = 0.5$, $1.5$ and $1.0$ respectively.

\subsection{Gravity}
\label{app:gravity}

GeoCalib~\citep{veicht2024geocalib} runs on 48 frames, and its camera-frame up vectors are lifted to the world with the VGGT-$\Omega$ rotations and averaged after inlier selection at $8^\circ$.
The estimate is accepted when the cameras span at least $8^\circ$ of rotation and the votes spread less than $5^\circ$.
Otherwise a RANSAC plane through the lowest fused points gives the down direction, and failing that the down axis of the first camera.
Only the direction is estimated.

\subsection{BEDLAM2}
\label{app:bedlam2_labels}

We import every BEDLAM2~\citep{bedlam2} sequence whose bodies all have a released motion in the AMASS format~\citep{mahmood2019amass}, excluding the egocentric and upper-body render jobs and the sequences flagged as test set, with the ground-truth SMPL-X~\citep{pavlakos2019smplx} bodies, cameras and the floor height of each sequence.
The shape is truncated to 10 coefficients, a frame is valid when the person is visible, and the person masks $\mathbf{s}_t$ that the backbone needs are rasterized from the ground-truth meshes.

With the ground-truth trajectory and cameras, no perception model runs and the contact rule of \appref{app:contacts} is replaced by a near-and-still rule against the floor plane of the sequence.
A point is in contact when it is below 8\,cm above the floor, leaving at 10\,cm, faces down to within $55^\circ$, stays for at least $0.1$\,s and moves slower than $0.2$\,m/s, the speed being the calmer of its means over the preceding and the following $0.1$\,s.
The forces are then solved at the frozen ground-truth pose with the dynamics terms of \eqref{eq:edyn} and \eqref{eq:ereg}, plus a ramp term that penalizes the force on the first and last frame of each contact, weakened by $1 / (1 + (v / 0.8\,\text{m/s})^2)$ with $v$ the vertical speed of the root, so that loads ramp in and out unless the body is moving fast.
The contact and force confidences are set to one.

\subsection{Benchmarks}
\label{app:label_benchmarks}

On \wall, the parkour sequences and OpenCap the camera is static, so the pipeline uses the scale of \eqref{eq:scale} without the refinement.
On \wall and the parkour sequences it drops the Chamfer term and restricts the contact points to the 18 of the hands and feet, and on the parkour footage the scene terms are disabled.
On OpenCap the scene is a plane fitted to the floor, with the chair and the box detected as plateaus, the contact points are the ten of the feet and the two sit bones, the ramp term of \appref{app:bedlam2_labels} is added, and the measured mass is given to the solver for the true-mass columns of \tabref{tab:optim_opencap}.

\section{Model and training details}
\label{app:model}

This appendix details the architecture of \secref{sec:architecture} and the losses of \secref{sec:losses}, following their notation.

\subsection{Contact-force tokens}
\label{app:tokens}

The $K = 35$ input embeddings $\mathbf{z}_t = (\mathbf{z}_{t,k})_{k=1}^{K}$ of \eqref{eq:contact_force_tokens} are learned vectors of width 1024, one per contact point of \figref{fig:contactpoints}, appended after the original token sequence of the six-layer SAM 3D Body decoder.
The attention mask is asymmetric: the original tokens do not attend to the appended ones, so the pose token $\mathbf{p}_t$ and the readout of the backbone are unchanged.
After each of the first five decoder layers, token $k$ adds a linear projection of the backbone features sampled on a $5\times5$ grid around the current projection of its point, located through the intermediate body readout of the decoder, together with a positional embedding of that projection.
The outputs $\mathbf{z}'_t$ are the appended tokens after the last layer.

\subsection{Body and camera heads}
\label{app:body_heads}

The heads $\mathcal{H}_{\mathrm{SMPL\text{-}X}}$ and $\mathcal{H}_{\mathrm{camera}}$ of \eqref{eq:body_configuration} are two feed-forward branches on the final pose token, of the same shape as the SAM 3D Body heads.
The first regresses the root and 21 body rotations and 30 finger rotations in 6D form and the 10 shape parameters, as residuals on a mean body, and the second a weak-perspective camera $\boldsymbol{\kappa}_t$ that CLIFF~\citep{li2022cliff} lifts to the full-frame translation $\mathbf{r}^c_t$ with the crop box and the focal length.

\subsection{World-informed pose token}
\label{app:pose_token}

The kinematics that enter $\operatorname{MLP}_{\mathrm{world}}$ in \eqref{eq:world_informed_features} are the joint positions and velocities of the world trajectory re-expressed in the body frame of frame $t$, so that nothing depends on the orientation of the world frame.
They comprise the joint positions and the joint rotations in 6D form, the backward and forward angular rates of the joints, the backward and forward linear and angular velocities of the root, the frame spacing, the clip-mean shape $\boldsymbol{\beta}$, and the direction, distance, bearing, size and axes of the camera, 351 channels in all.
In practice $\operatorname{MLP}_{\mathrm{world}}$ is a layer normalization of these channels and $\operatorname{MLP}_{\mathrm{pose}}$ a linear projection of $\mathbf{p}_t$ to 256 channels, and their concatenation $\mathbf{p}'_t$ is projected to width 512 by one linear layer.
The contact-force tokens $\mathbf{z}'_{t,k}$ are projected to the same width, and a learned embedding tells the $K + 1$ slots apart.

\subsection{Temporal transformer and iterative refinement}
\label{app:temporal}

The transformer $\mathcal{T}$ of \eqref{eq:temporal_aggregation} has three pre-norm blocks of width 512 with 8 heads, a feed-forward ratio of 4 and dropout $0.1$.
In each block, attention runs first along time within every slot, with rotary embeddings~\citep{su2024rope} on elapsed time and masked to $\pm0.083$\,s, then across the $K + 1$ slots within a frame, followed by the feed-forward layer.
Three blocks give a receptive field of $\pm0.25$\,s, so the model trained on 120-frame windows is applied to whole videos, resampled to about 25\,fps, in windows of 240 frames overlapping by 120.

The three blocks $\mathcal{T}_1, \mathcal{T}_2, \mathcal{T}_3$ are applied one at a time, and the pose is corrected after each of them, which is the iterative refinement of \secref{sec:prediction}.
With $(\mathbf{p}^{(0)}_t, \mathbf{z}^{(0)}_t) = (\mathbf{p}'_t, \mathbf{z}'_t)$ and $\hat{\mathbf{q}}^{(0)}_t = \mathbf{q}_t$, block $i$ computes
\begin{equation}
    (\mathbf{p}^{(i)}_t, \mathbf{z}^{(i)}_t)_{t=1}^{T} = \mathcal{T}_i\big( (\mathbf{p}^{(i-1)}_t, \mathbf{z}^{(i-1)}_t)_{t=1}^{T} \big),
    \qquad
    \hat{\mathbf{q}}^{(i)}_t = \hat{\mathbf{q}}^{(i-1)}_t \oplus \mathcal{H}_{\mathrm{offset}}(\mathbf{p}^{(i)}_t),
    \label{eq:iteration}
\end{equation}
after which the body model is re-run in the world with the clip-mean shape and the contact, force and gravity heads are read from $(\mathbf{p}^{(i)}_t, \mathbf{z}^{(i)}_t)$.
The refresh of \eqref{eq:world_informed_features} before the next block is a residual update of the pose token with the recomputed kinematics and the current predictions,
\begin{equation}
    \mathbf{p}^{(i)}_t \leftarrow \mathbf{p}^{(i)}_t + \mathbf{W}_{\text{fb}}\, \operatorname{LN}\big( \mathbf{j}_t(\hat{\mathbf{q}}^{(i)}),\, \dot{\mathbf{j}}_t(\hat{\mathbf{q}}^{(i)}),\, \hat{\mathbf{c}}^{(i)}_t,\, \mathbf{R}_t^{\top}\hat{\mathbf{g}}^{(i)} \big),
    \label{eq:feedback}
\end{equation}
with $\mathbf{W}_{\text{fb}}$ a zero-initialized linear layer and $\hat{\mathbf{g}}^{(i)}$ the clip gravity estimate of \appref{app:heads}, rotated into the body frame of frame $t$ by the root orientation $\mathbf{R}_t$.
The outputs of \eqref{eq:temporal_aggregation} are those of the last block, $(\mathbf{p}''_t, \mathbf{z}''_t) = (\mathbf{p}^{(3)}_t, \mathbf{z}^{(3)}_t)$ and $\hat{\mathbf{q}}_t = \hat{\mathbf{q}}^{(3)}_t$, and the contact, force, trajectory and gravity losses are also applied to the predictions of the first two blocks at weight $0.5$.
The trainable part has 19.7M parameters on top of the 1.3B frozen backbone and decoder.

\subsection{Prediction heads}
\label{app:heads}

The heads $\mathcal{H}_{\mathrm{contacts}}$, $\mathcal{H}_{\mathrm{force}}$ and $\mathcal{H}_{\mathrm{offset}}$ of \eqref{eq:contact_force_prediction} and \eqref{eq:pose_correction} are two-layer MLPs with GELU, shared across the three blocks.
The offset head yields 6D corrections of the root and the 21 body joints and a root shift in the body frame, and the finger rotations are left as predicted per frame.
The force head regresses $\tilde{\mathbf{f}}_{t,k}$ in the body frame of frame $t$ and in units of body weight, and the gated world force in newtons of \eqref{eq:contact_force_prediction} is
\begin{equation}
    \hat{\mathbf{f}}_{t,k} = \hat{c}_{t,k}\, mg\, \mathbf{R}_t\, \tilde{\mathbf{f}}_{t,k},
    \label{eq:force_world}
\end{equation}
with $\mathbf{R}_t$ the root orientation of the input trajectory $\mathbf{q}_t$ and $m$ the mass of the reconstructed body.
The gravity head of \eqref{eq:pose_correction} predicts from $\mathbf{p}''_t$ a per-frame vote for the down axis, as a correction of the down axis of the camera, and the unit votes are averaged over the clip and renormalized into the gravity direction $\hat{\mathbf{g}}$.

\subsection{Losses}
\label{app:losses}

\Tabref{tab:losses} lists the terms and weights.
The labels are the trajectory $\mathbf{q}^*$, the contacts $c^*_{t,k}$ and the forces $\mathbf{f}^*_{t,k}$ of \eqref{eq:edyn}, with the confidences $w^{\text{c}}_{t,k}$ and $w^{\text{f}}_t$ of \appref{app:confidences}, and $\rho$ denotes the Huber loss of \eqref{eq:lforce} with the scale given in \tabref{tab:losses}.

\begin{table}[h]
\caption{\textbf{Training losses.} Terms and weights. Confidence-weighted terms are marked in the last column, and the contact, force, trajectory and gravity terms also supervise the intermediate blocks at weight 0.5. Huber losses are given with their scale.}
\label{tab:losses}
\begin{center}
\small
\resizebox{\linewidth}{!}{%
\begin{tabular}{llrc}
\toprule
Loss & Definition & Weight & Conf. \\
\midrule
\multicolumn{4}{l}{\emph{Per-frame body $\mathcal{L}_{\text{body}}$ (SMPL-X and camera heads)}} \\
2D joints & Huber on the full-frame projection, crop units, $\delta = 0.05$ & 10 & \\
3D joints & Huber, pelvis-relative, $\delta = 0.1$\,m & 5 & \\
Root, body and finger rotations & squared error on 6D & 1, 1, 0.5 & \\
Shape & squared error & 0.1 & \\
Camera & Huber on the weak-perspective parameters, $\delta = 0.05$ & 1 & \\
\midrule
\multicolumn{4}{l}{\emph{Corrected trajectory $\mathcal{L}_{\text{traj}}$ (offset head)}} \\
3D joints, rotations & as above & 10, 2 & \\
World root & clip mean ($\delta = 0.1$\,m), per-frame deviation ($\delta = 0.02$\,m) & 2, 2 & \\
Joint velocity, root angular velocity & Huber on forward differences, scaled by 0.4\,m/s and 0.57\,rad/s & 1, 1 & \\
\midrule
Contacts $\mathcal{L}_{\text{c}}$ & binary cross-entropy, per point & 35/6 & $\checkmark$ \\
Force vector $\mathcal{L}_{\text{f}}$, magnitude $\mathcal{L}_{\text{mag}}$ & Huber, $\delta = 0.5$\,bw, on labeled contacts & 1, 1 & $\checkmark$ \\
Force off contact (in $\mathcal{L}_{\text{f}}$) & $L_1$ on the magnitude & 1 & $\checkmark$ \\
Net force $\mathcal{L}_{\text{net}}$ & Huber on the sum over points & 1 & $\checkmark$ \\
Contact stillness $\mathcal{L}_{\text{still}}$ & $L_1$ on the world speed of a labeled contact & 1 & $\checkmark$ \\
Gravity $\mathcal{L}_{\text{g}}$ & $1 - \cos$, scenes with an accepted gravity estimate & 5 & \\
Physics $\mathcal{L}_{\text{phys}}$ & $\lambda_F = 5$, $\delta_F = 0.2$; $\lambda_T = 20$, $\delta_T = 0.5$; $\lambda_J = 0.3$ & & \\
\bottomrule
\end{tabular}}
\end{center}
\end{table}

\paragraph{Per-frame body.}
The heads of \appref{app:body_heads} are trained as in HMR 2.0~\citep{hmr2} against the labels projected into the camera, with Huber losses on the full-frame projection of the joints, on the pelvis-relative 3D joints and on the weak-perspective camera, squared errors on the rotations in 6D form and on the shape, and the finger joints weighted $0.1$.

\paragraph{Corrected trajectory.}
The corrected trajectory $\hat{\mathbf{q}}$ of \eqref{eq:pose_correction} is supervised on its pelvis-relative joints and rotations, on its world root and on the velocities of its joints and of its root orientation:
\begin{equation}
    \begin{aligned}
    \mathcal{L}_{\text{traj}} = \sum_t \Big[ &
        \lambda_{\text{j}}\, \rho\big( \mathbf{j}_t(\hat{\mathbf{q}}) - \mathbf{j}_t(\mathbf{q}^*) \big)
        + \lambda_{\theta}' \big\| \hat{\boldsymbol{\theta}}_t - \boldsymbol{\theta}^*_t \big\|^2 \\
        & + \lambda_{\text{v}}\, \rho\Big( \frac{\Delta \mathbf{j}_t(\hat{\mathbf{q}}) - \Delta \mathbf{j}_t(\mathbf{q}^*)}{\Delta t} \Big)
        + \lambda_{\omega}\, \rho\Big( \frac{\Delta \hat{\boldsymbol{\theta}}_{t,0} - \Delta \boldsymbol{\theta}^*_{t,0}}{\Delta t} \Big)
    \Big] \\
    & + \lambda_{\text{r}}\, \rho\big( \bar{\mathbf{e}} \big)
        + \lambda_{\text{r}}' \sum_t \rho\big( \hat{\mathbf{r}}_t - \mathbf{r}^*_t - \bar{\mathbf{e}} \big),
    \qquad
    \bar{\mathbf{e}} = \frac{1}{T} \sum_t \big( \hat{\mathbf{r}}_t - \mathbf{r}^*_t \big),
    \end{aligned}
    \label{eq:ltraj}
\end{equation}
where the rotations of the body joints are compared in 6D form and the differences of the root orientation are taken in $\mathfrak{so}(3)$.
The root translation is scored on the clip-mean error $\bar{\mathbf{e}}$ and on the per-frame deviation from it, since the scale of a monocular reconstruction is uncertain as a whole.

\paragraph{Contacts and forces.}
The contact loss $\mathcal{L}_{\text{c}}$ of \eqref{eq:lforce} is normalized per point before summing, so that the rare points weigh as much as the frequent ones.
The forces $\tilde{\mathbf{f}}_{t,k}$ and $\mathbf{f}^*_{t,k}$ of \eqref{eq:lforce} are compared in the body frame of the input trajectory $\mathbf{q}_t$, into which the label is rotated, in units of body weight, and $\mathcal{L}_{\text{f}}$ is complemented by a term on the magnitude of each labeled force and by one on the net force over the points,
\begin{equation}
    \mathcal{L}_{\text{mag}} = \sum_{t,k} w^{\text{f}}_t\, \bar{c}^*_{t,k}\, \rho\big( \| \tilde{\mathbf{f}}_{t,k} \| - \| \mathbf{f}^*_{t,k} \| \big),
    \qquad
    \mathcal{L}_{\text{net}} = \sum_t w^{\text{f}}_t\, \rho\Big( \sum_k \tilde{\mathbf{f}}_{t,k} - \sum_k \mathbf{f}^*_{t,k} \Big),
    \label{eq:lmagnet}
\end{equation}
which ties the sum of the predictions to the sum of the labels even when the split between points is uncertain.
Contact points whose labeled force exceeds four body weights, a sign of a wrong scale or camera, are excluded from the force losses.

\paragraph{Contact stillness.}
A point labeled in contact should not move in the world, so the corrected trajectory is penalized on the speed of its labeled contacts,
\begin{equation}
    \mathcal{L}_{\text{still}} = \sum_{t,k} w^{\text{c}}_{t,k}\, c^*_{t,k}\, \frac{\| \Delta \mathbf{v}_{t,k}(\hat{\mathbf{q}}) \|}{\Delta t},
    \label{eq:lstill}
\end{equation}
with $\mathbf{v}_{t,k}(\hat{\mathbf{q}})$ the world position of contact point $k$ on the corrected body.

\paragraph{Gravity.}
With $\mathbf{g}^*$ the scene gravity of \appref{app:gravity}, $\mathcal{L}_{\text{g}} = 1 - \hat{\mathbf{g}}^{\top} \mathbf{g}^*$ per clip on the scenes whose GeoCalib estimate is accepted.

\subsection{Physics loss}
\label{app:physics_loss}

The physics loss of \eqref{eq:lphys} evaluates \eqref{eq:rnea} on the corrected trajectory $\hat{\mathbf{q}}$ with the predicted forces.
The body is the multibody of \appref{app:dynamics} reduced to the 22 body joints, with the finger parts merged into the hand segments and masses and inertias integrated from the clip-mean shaped mesh, the velocities and accelerations are central differences on the manifold with a $\pm2$-frame stencil for the acceleration, and gravity is the estimate $\mathbf{g}^*$ of the scene (\appref{app:gravity}).
Each predicted force enters in world newtons as in \eqref{eq:force_world} with its gate detached,
\begin{equation}
    \begin{aligned}
    \hat{\mathbf{f}}_{t,k} &= \operatorname{sg}\big( \hat{c}_{t,k} \big)\, mg\, \mathbf{R}_t\, \tilde{\mathbf{f}}_{t,k}, \\
    \boldsymbol{\tau}_t &= \mathbf{M}(\hat{\mathbf{q}}_t)\, \ddot{\hat{\mathbf{q}}}_t + \mathbf{b}(\hat{\mathbf{q}}_t, \dot{\hat{\mathbf{q}}}_t) + \mathbf{G}(\hat{\mathbf{q}}_t) - \mathbf{J}(\hat{\mathbf{q}}_t)^{\top} \hat{\mathbf{f}}_t,
    \end{aligned}
    \label{eq:lphys_rnea}
\end{equation}
where $\operatorname{sg}$ is the stop-gradient and $\mathbf{J}(\hat{\mathbf{q}}_t)$ the Jacobian of the contact points, so that each force acts at its point $\mathbf{v}_{t,k}(\hat{\mathbf{q}})$ and produces a lever moment on its parent joint.
Detaching the gate keeps the contact probabilities from being driven by the physics, which would let the model switch a contact off to cancel a residual.
The root force and moment $\boldsymbol{\tau}^{f}_t$ and $\boldsymbol{\tau}^{m}_t$ of the result are the residual that the forces fail to explain and enter \eqref{eq:lphys} through the pseudo-Huber loss
\begin{equation}
    \rho_\delta(\mathbf{e}) = \frac{1}{3} \sum_{i=1}^{3} \delta^2 \Big( \sqrt{1 + e_i^2 / \delta^2} - 1 \Big),
    \label{eq:pseudohuber}
\end{equation}
applied per component, which is quadratic on small residuals and linear on the large ones that a wrong label or a missed contact produces.
The joint-torque term of \eqref{eq:lphys} uses the multipliers $\mu_j$ of \tabref{tab:torque} without the fingers, which the corrected trajectory does not move, and acts on the forces only, the body being detached in that term, so that it distributes the load between limbs without pulling on the pose.
The loss is averaged over the frames whose stencil is valid and backpropagates into the force head and, through the root terms, into the offset head.

\subsection{Training}
\label{app:training}

Clips are 120 frames, resampled to about 25\,fps for \datasetC with a random offset per epoch and at 30\,fps for BEDLAM2, with no other augmentation.
All runs use AdamW with weight decay $0.01$, a cosine schedule to $10^{-6}$ and an exponential moving average of the weights, which is what we evaluate.
The \datasetC model trains for 10 epochs at learning rate $3\times10^{-4}$ with 300 warm-up steps and 32 clips per step, and the BEDLAM2 model with the same schedule on a fresh random third of the sequences per epoch with 24 clips per step.
The joint model starts from the BEDLAM2 model and trains for 5 epochs at learning rate $10^{-4}$ without warm-up on 16 \datasetC and 16 BEDLAM2 clips per step, one pass over \datasetC per epoch and a fresh 7\% of BEDLAM2.
Each ablation rung of \tabref{tab:ablation} is trained from scratch with the \datasetC schedule, and the last checkpoint of every run is reported.

\section{Datasets and metrics}
\label{app:curation}
\label{app:metrics}

\Tabref{tab:datasets} summarizes the data and \tabref{tab:funnel} the curation of \datasetC.
Contacts are scored on every dataset by precision, recall and F1 at threshold $0.5$, without temporal tolerance, micro-averaged over frames and limbs.
The $K$ predicted probabilities are folded to a limb $b$ by their maximum, $\hat{c}_{t,b} = \max_{k \in b} \hat{c}_{t,k}$.
Per-vertex estimators are folded to the limbs by the skinning weights or the part segmentation of the vertices, either any contact vertex or a tenth of them, and the best of the four rules is reported on \datasetC and \wall, while on RICH the predictions are folded by the rule of its labels (\appref{app:rich}).
GVHMR and PromptHMR report their stationary-joint probabilities at $0.5$.

\begin{table}[h]
\caption{\textbf{Datasets.} Data used in this paper.}
\label{tab:datasets}
\begin{center}
\small
\resizebox{\linewidth}{!}{%
\begin{tabular}{lrrrl}
\toprule
 & Sources & Scenes & Frames & Labels \\
\midrule
\datasetC train & 149 & 1{,}419 & 460k & pseudo \\
\datasetC test & 42 & 109 & 31k & pseudo, manual contacts \\
BEDLAM2~\citep{bedlam2} & 59 jobs & 18{,}076 & 5.5M & true pose, pseudo contacts and forces \\
\wall & 14 & 68 & 29.6k & sensors, multi-view pose \\
RICH~\citep{huang2022rich} & 8 subjects & 164 & 70.2k & contact labels \\
Parkour~\citep{li2019_forces} & -- & 28 & 2.3k & motion capture, plates and bar \\
OpenCap~\citep{uhlrich2023opencap} & 9 subjects & 64 & -- & force plates \\
\bottomrule
\end{tabular}}
\end{center}
\end{table}

\begin{table}[h]
\caption{\textbf{Curation funnel.} Curation of \datasetC. Counts are YouTube videos until the scene split and scenes afterwards.}
\label{tab:funnel}
\begin{center}
\small
\begin{tabular}{p{0.82\linewidth}r}
\toprule
Stage & Count \\
\midrule
Videos returned by keyword search (``bouldering gym'', ``moonboard climbing'' and the like) & 782 \\
Duration between 10\,s and 20\,min & 591 \\
Thumbnail and metadata VLM gate & 584 \\
Downloaded at up to 1080p (48\,h of video) & 581 \\
\midrule
Scenes after shot detection, between 100 and 500 frames & 16{,}152 \\
Full-body pose gate: in half of the sampled frames, a person with 60\% of the 17 COCO keypoints visible, including an ankle, a shoulder and a hip, spanning 40\% of the frame height & 4{,}612 \\
Video VLM gate: moving climber, single real-time shot & 3{,}738 \\
Indoor and not rope-supported & 3{,}071 \\
Manual selection (from 191 videos) & 1{,}528 \\
\bottomrule
\end{tabular}
\end{center}
\end{table}

\subsection{\datasetC}
\label{app:climbingvideos}

\paragraph{Curation.}
The gates and their thresholds are listed in \tabref{tab:funnel}.
Shots are detected with a content-change detector, long scenes are split uniformly, and a boundary guard removes frames leaking from a neighboring shot.
The full-body gate runs a YOLO pose detector~\citep{jocher2023ultralytics} on every fifth frame, and the video VLM~\citep{clark2026molmo2} answers a fixed rubric, including the type of climbing and whether a rope carries the climber's weight, on 24 uniformly sampled frames.
The climber it points at is associated with a track of BoT-SORT~\citep{aharon2022bot} on top of the pose detector.

\begin{figure}[t]
\begin{center}
\includegraphics[width=0.85\linewidth]{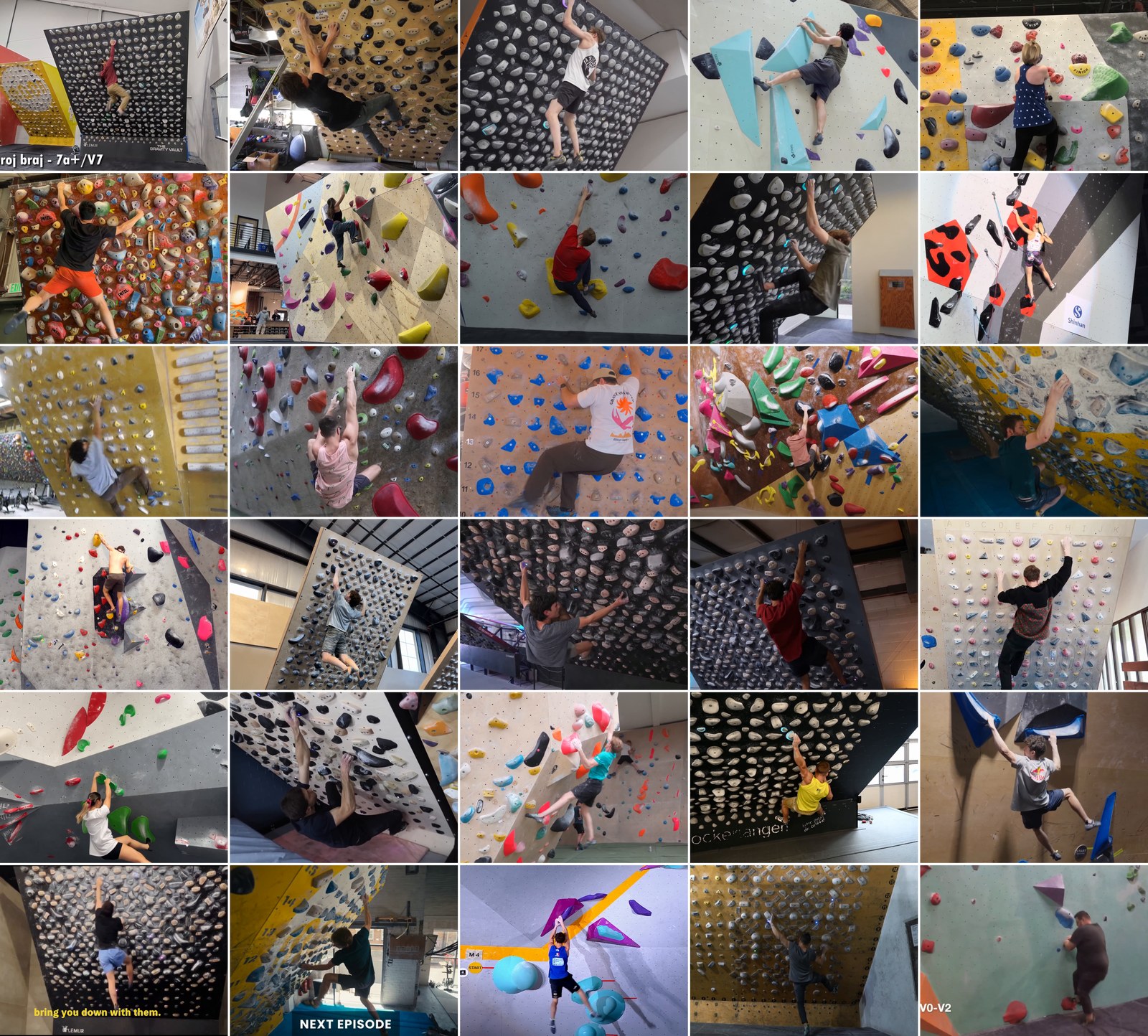}
\end{center}
\caption{\textbf{\datasetC samples.} One frame from each of 30 scenes, spanning gyms and competition walls, slabs to steep overhangs, static and moving cameras, and varied lighting and clothing.}
\label{fig:dataset}
\end{figure}

\paragraph{Statistics.}
The selected scenes (\figref{fig:dataset}) have a mean length of 322 frames (10.7\,s), 92\% are 1080p, and the source frame rate is 24 to 60\,fps with a median of 25.
The split is by source video, with the test scenes drawn from videos with at most four scenes.

\begin{figure}[t]
\begin{center}
\includegraphics[width=0.75\linewidth]{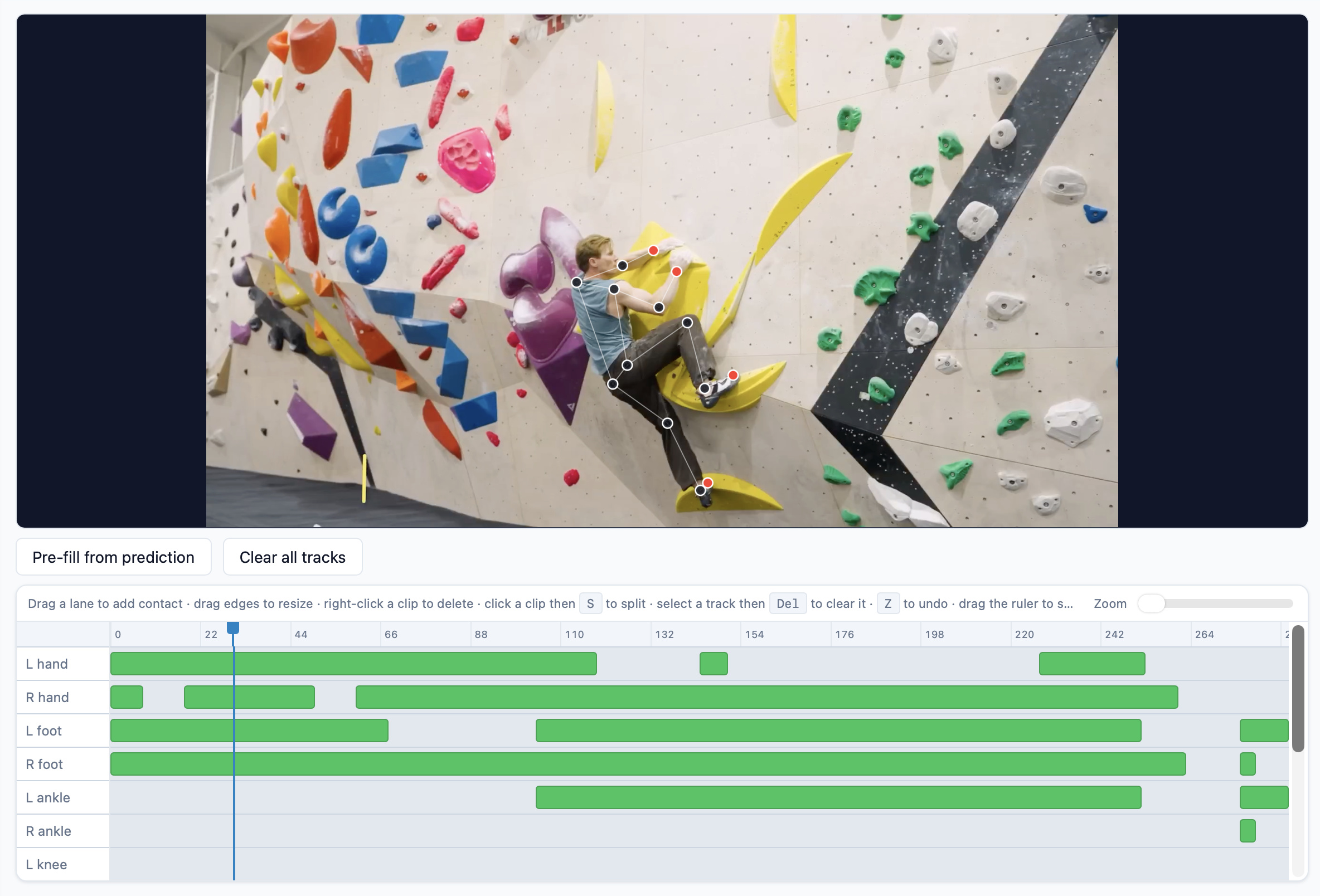}
\end{center}
\caption{\textbf{Contact annotation tool.} The interface used to annotate contacts on the \datasetC test split.}
\label{fig:annotator}
\end{figure}

\paragraph{Contact labels and scoring.}
The test split carries a contact label (in contact, free or unknown) for 14 joints on its 31k frames, made with the tool of \figref{fig:annotator}.
Hands are in contact in 84\% of the frames, toes in 63\%, heels in 8\%, and knees, elbows, shoulders and hips in less than 2\%.
Contacts are scored on six limbs, the two hands, the two toes and the two heels, to which the annotated hand, toe and heel labels are folded, and the 17 points outside them are not scored.

\subsection{\wall}
\label{app:forcewall}

\begin{figure}[t]
\begin{center}
\begin{subfigure}[b]{0.42\linewidth}
\centering
\includegraphics[height=6.5cm]{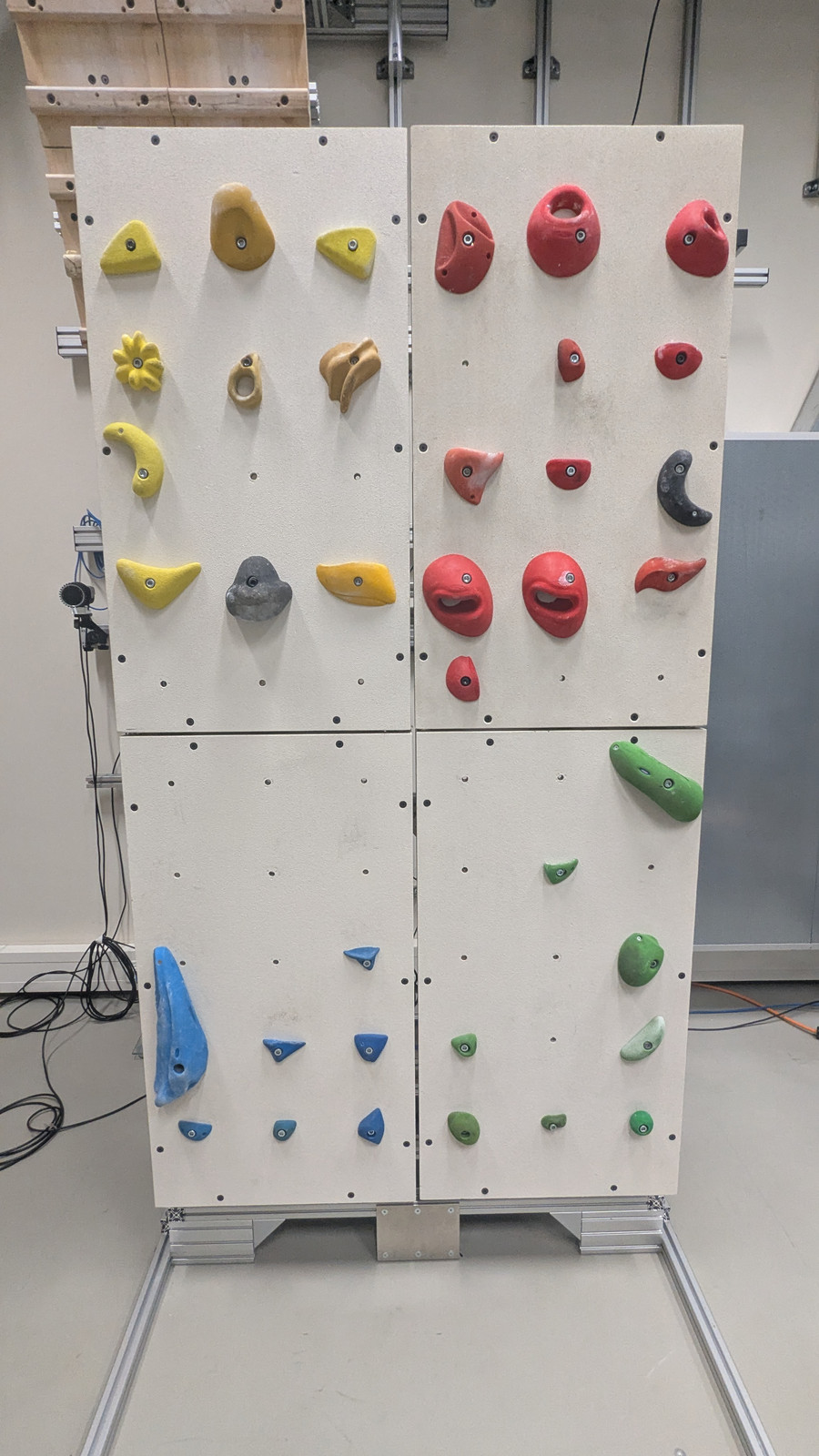}
\caption{Front view}
\end{subfigure}
\hfill
\begin{subfigure}[b]{0.42\linewidth}
\centering
\includegraphics[height=6.5cm]{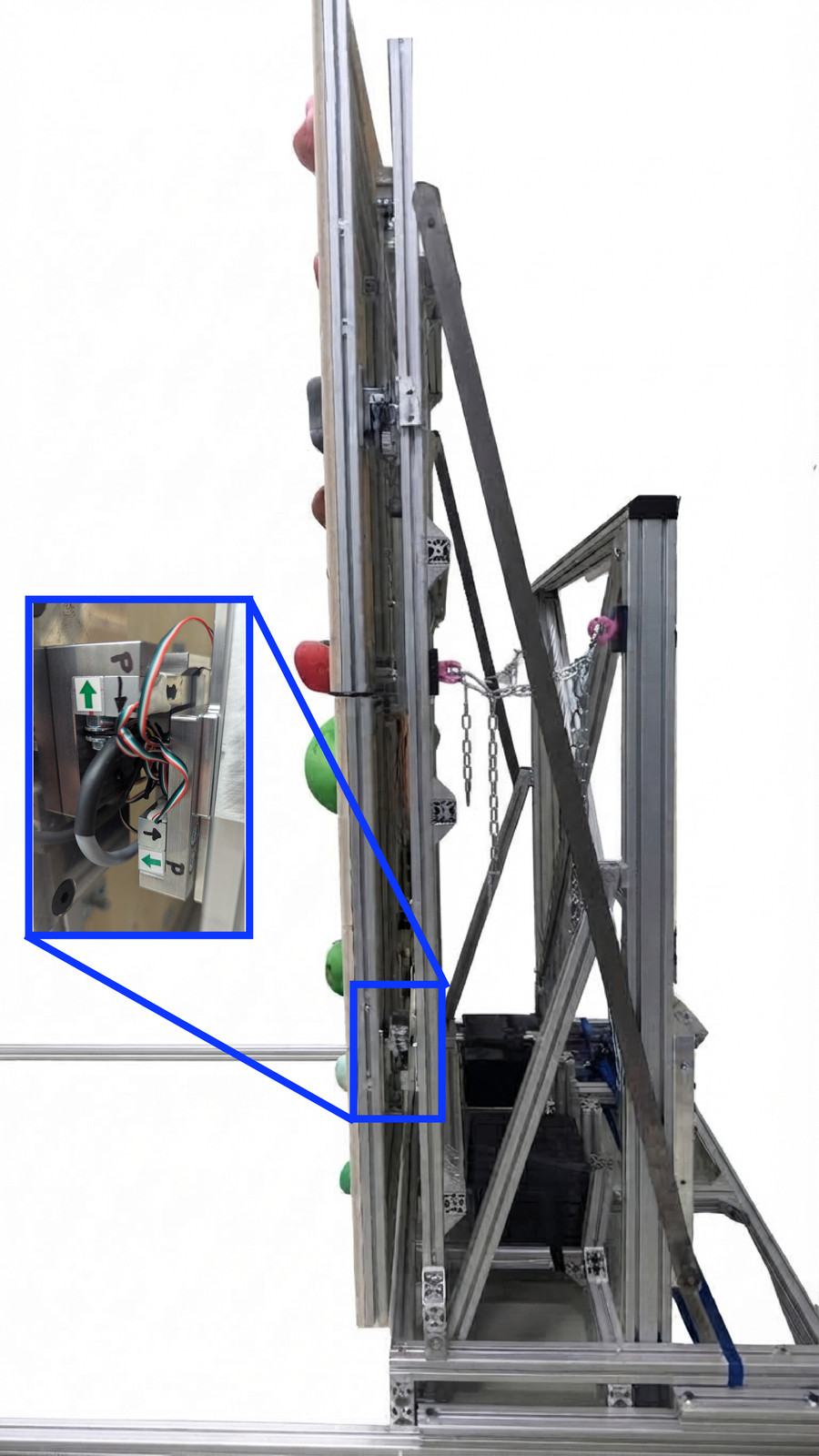}
\caption{Side view}
\end{subfigure}
\end{center}
\caption{\textbf{\wall hardware.} (a) The four instrumented panels with their holds. (b) The tilted frame that carries the panels, with a zoom on one of the three-axis load cells between a panel and the frame.}
\label{fig:wall_setup}
\end{figure}

\paragraph{Hardware.}
The wall (\figref{fig:wall_setup}) has four instrumented panels of 0.6$\times$1.1\,m, the quarters of a bouldering board, so that in normal climbing each limb loads one panel and a sensor channel is the force of one limb.
Each panel rests on three custom three-axis load cells, each built from six single-point load cells (CZL635, Phidgets), sampled at 100\,Hz through an amplifier.
The panels were calibrated along each axis with known loads, with an accuracy of at least 95\% and a crosstalk of at most 5\%.
The frame is tilted to $0^\circ$, $16^\circ$ or $32^\circ$, and the holds mounted on the panels elicit side pulls, underclings, pockets and crimps for the hands as well as heel hooks, knee drops and flagging.

\paragraph{Recording.}
Two participants, one male (72.5\,kg, 1.80\,m) and one female (48\,kg, 1.60\,m), perform bouldering movements in a self-chosen order, recorded at 30\,fps and 1080p by three webcams (C920, Logitech), two behind the participant at different angles and one lateral, and cut into clips of at most 500 frames.
Forces are debiased on a quiet stretch and mapped to the world with the sensor datasheet and the wall's azimuth from the reconstructed scene.
The sensors and the three cameras are synchronized on the first loading of the left hand and on the reprojection error of the stereo fit.
Ground-truth contact is a load above 50\,N entering and 30\,N leaving, scored on the four limbs, a foot being in contact when its toes or its heel are.

\paragraph{Reference pose.}
The multi-view reference pose is fitted with the optimization of \appref{app:autolabel} with three changes.
The reprojection terms act in all three cameras, calibrated with a chessboard (0.2 to 0.3\,px reprojection error), the SMPL-X shape of each climber is fitted once against the true height and weight and then frozen, and the measured contacts and forces enter as constraints, the latter through a term that pulls the sum of the point forces of each limb to its sensor.
The fit is seeded in every view by SAM 3D Body, and 29.6k of its frames are valid and scored.
Pose is scored by MPJPE on the 22 body joints after hip centering in the camera frame.

\paragraph{Force metrics.}
\label{app:metrics_forces}
The predicted forces are summed by limb $b$, $\hat{\mathbf{F}}_{t,b} = \sum_{k \in b} \hat{\mathbf{f}}_{t,k}$, in newtons with the mass $m$ that our pipeline reconstructs for the clip (the baselines keep the mass of their own body model, \appref{app:baselines}).
They are compared with the sensor force $\mathbf{F}_{t,b}$ on the set $\mathcal{S}$ of limb-frames $(t, b)$ with a valid reference, in contact or not, so that force leaked onto a free limb is penalized:
\begin{equation}
    \text{MAE} = \frac{1}{|\mathcal{S}|} \sum_{(t,b) \in \mathcal{S}} \big\| \hat{\mathbf{F}}_{t,b} - \mathbf{F}_{t,b} \big\|.
    \label{eq:mae}
\end{equation}
The angle is the mean of $\arccos\big( \hat{\mathbf{F}}_{t,b}^{\top} \mathbf{F}_{t,b} / (\| \hat{\mathbf{F}}_{t,b} \| \| \mathbf{F}_{t,b} \|) \big)$ over the limb-frames where both magnitudes exceed 50\,N, and the correlation is the Pearson correlation of the magnitudes $\| \hat{\mathbf{F}}_{t,b} \|$ and $\| \mathbf{F}_{t,b} \|$ pooled over frames and limbs.
The load-share error is the mean absolute difference, in percentage points, between the predicted share $\| \hat{\mathbf{F}}_{t,b} \| / \sum_{b'} \| \hat{\mathbf{F}}_{t,b'} \|$ of a limb and the measured share $\| \mathbf{F}_{t,b} \| / \sum_{b'} \| \mathbf{F}_{t,b'} \|$.
The physical residual is the mean of $\| \boldsymbol{\tau}^{f}_t \| / (mg)$ and of $\| \boldsymbol{\tau}^{m}_t \| / (mg)$ with the root wrench of \eqref{eq:lphys_rnea}, that is, with the forces gated by the contact probabilities, in body weights (bw) and body weights times meters.

\subsection{RICH}
\label{app:rich}

We score the static-camera sequences of the RICH~\citep{huang2022rich} test split on the frames of the contact evaluation of BSTRO, excluding the two-person sequences: 164 clips, one per sequence, camera and subject, with 70{,}159 frames of 8 subjects in three scenes.
The per-vertex labels of RICH are folded to the six limbs of \appref{app:climbingvideos} through the SMPL joint with the largest skinning weight of each vertex, the hands from the wrist and hand joints, the toes from the foot joint and the heels from the ankle joint, and a limb is in contact when at least one of its vertices is labeled.
Every method receives the ground-truth person box, our model and PromptHMR also the ground-truth silhouette as the mask, and the video models run on the full timeline of each sequence and are scored on the annotated frames only.
The per-vertex predictions of BSTRO, DECO and InteractVLM are thresholded at $0.5$ and folded by the same rule.

\subsection{Parkour}
\label{app:parkour_data}

The 28 sequences of \citet{li2019_forces} are archival PAL footage (720$\times$576, 25\,fps) with Vicon motion capture on 16 joints and contact forces measured by force plates under the feet and an instrumented bar under the hands, 2{,}324 reference frames in all.
We score every method with the evaluator shipped with the dataset, against the original force files of the authors, and convert every method's forces to newtons with the 74.6\,kg body that the evaluator assumes, so that no method is favored by its mass estimate.
The evaluator aligns each frame to the 16 Vicon joints by a similarity transform and reports the mean joint error, and scores the summed force of each limb against the plates and the bar by the mean norm of the vector difference over the frames of a clip, averaged over clips.

\subsection{OpenCap}
\label{app:opencap_data}

The validation set of \citet{uhlrich2023opencap} was recorded with smartphones on nine subjects walking, squatting, rising from a chair and dropping from a box over force plates.
We take one camera and the natural-speed trials, exclude six whose synchronization with the plates we could not verify, and score the remaining 64, of which 22 walking trials of eight subjects form the walking subset.
The forces of every method and the plates are filtered with a fourth-order zero-lag Butterworth filter, at 6\,Hz for walking and 4\,Hz for squats and sit-to-stand, with the drop jumps unfiltered.
The vertical, anteroposterior and mediolateral errors of each foot, in percent of body weight, are averaged over the frames of a trial, then over the trials of a subject, then over subjects.
Over the four activities every frame on which the plates watched a foot counts, the protocol of \citet{uhlrich2023opencap}, and on walking only the stance frames, where the plate reads more than 5\% of body weight, the protocol of \citet{gilon2026opencapmono}, whose numbers are taken from their paper.
Every method that estimates its own mass is scored twice, with its forces in its own newtons and with them rescaled by the ratio of the measured to its estimated mass, which gives the true-mass columns of \tabref{tab:opencap} and \tabref{tab:optim_opencap}, except for our optimization, where the measured mass is given to the solver.

\subsection{Baselines}
\label{app:baselines}

\paragraph{\citet{li2019_forces}++.}
We run the released estimator of \citet{li2019_forces} with our inputs in place of its own: the per-frame SAM 3D Body body as the initial pose, with the SMPL shape fitted to its skeleton, the Sapiens2 keypoints folded to the OpenPose joints, and its own contact recognizer applied to those keypoints.
On \wall the method also receives the calibrated camera and the gravity direction of our pipeline, fits its ground plane through the feet in contact and frees the plane in its last stage, and the hands hold a virtual object as in the original method.

\paragraph{PhysPT.}
We run the released PhysPT~\citep{physpt} with its checkpoints on the per-frame SAM 3D Body poses in place of its CLIFF input, without translation, which PhysPT predicts itself.
Its per-vertex ground reaction forces are summed per foot and placed at their force-weighted centroid, a foot counts as in contact above 20\,N, and the hands never carry force.
PhysPT leaves the last frame of every clip unpredicted, which is excluded on \wall and counts as zero force on the parkour sequences.

\paragraph{Inverse dynamics.}
The per-frame SAM 3D Body poses are smoothed with a Gaussian of $0.1$\,s, and the contacts are the InteractVLM (``wall'') vertex probabilities folded to the limbs, a limb being in contact when a tenth of its vertices exceed the threshold.
One force per limb in contact, acting at the wrist or at the base of the toes, solves the six root-wrench equations of \eqref{eq:rnea} per frame in the least-squares sense, without friction cone, joint torque or smoothness.

\paragraph{Units.}
On \wall our forces and those of the inverse dynamics are converted to newtons with the mass that our pipeline reconstructs for the clip, \citet{li2019_forces}++ and PhysPT keep the newtons of their own body models, and on the parkour sequences every method is rescaled to the 74.6\,kg body of the evaluator (\appref{app:parkour_data}).

\section{Limitations and future work}
\label{app:limitations}

\ours models the interaction between a person and a rigid, static scene, with three simplifications.
First, hand-held objects are not represented, so neither the pipeline nor the model accounts for their forces, which leaves lifting, carrying and tool use out of scope. Retrieving the objects and adding their dynamics to the optimization of \secref{sec:training_data} would extend the labels, and the model, to these interactions.
Second, every support is assumed rigid, so mats, ropes or trampolines, whose forces depend on their deformation, are out of reach, and the method is restricted to activities on rigid ground and structures.
Third, the body is the articulated rigid body of SMPL-X with the segments of \appref{app:dynamics}, and we estimate contact forces rather than the muscle forces that produce them. Muscle estimation is a separate problem that requires a musculoskeletal body, for which a biomechanically accurate skeleton such as SKEL~\citep{keller2023skel} is a natural step, although few muscled models adapt to the shape of each subject.

\section{Additional results}
\label{app:extra}

\subsection{Ablation rungs}
\label{app:ablation}

The rungs of \tabref{tab:ablation} are cumulative.
The per-frame model reads contacts, forces and gravity from the frozen pose token $\mathbf{p}_t$ of each frame with the SMPL-X and camera heads and nothing else.
The temporal transformer rung adds three blocks of attention over the pose tokens along time, without the world lift of \eqref{eq:world_informed_features} or any temporal loss.
The world-informed rung replaces them with the temporal transformer of \secref{sec:prediction} over the pose token $\mathbf{p}'_t$, with the lift to the world, the camera context, the offset head and the losses $\mathcal{L}_{\text{traj}}$ and $\mathcal{L}_{\text{still}}$, in a single pass.
The next rungs add the iteration of \eqref{eq:iteration} with the feedback of \eqref{eq:feedback}, the $K$ contact-force tokens $\mathbf{z}_t$ in the decoder and the transformer, and $\mathcal{L}_{\text{phys}}$ with its root-wrench and joint-torque terms.
The model without a contact head keeps the $K$ tokens and the force head, drops $\mathcal{L}_{\text{c}}$ and $\mathcal{L}_{\text{still}}$ and applies $\mathcal{L}_{\text{phys}}$ to the ungated forces.

\subsection{Temporal attention window}
\label{app:window}

\Tabref{tab:window} varies the temporal mask of every block on the model trained on \datasetC, from half of the default $\pm0.083$\,s to no mask, so that the receptive field of the three blocks goes from $\pm0.125$\,s to the whole 240-frame inference window.
The force error moves by about 2\,N over the range and the other metrics by a similar margin, the widest window being slightly better on correlation and load share and slightly worse on pose.
The window therefore matters little beyond a few tenths of a second, and the default keeps the model applicable to clips of any length at a fixed cost per frame.

\begin{table}[h]
\caption{\textbf{Temporal attention window.} On \wall, with the model trained on \datasetC and the protocol of \tabref{tab:ablation}. The window is the receptive field of the three blocks together.}
\label{tab:window}
\centering
\small
\setlength{\tabcolsep}{3pt}
\resizebox{\linewidth}{!}{%
\begin{tabular}{l*{8}{c}}
\toprule
\multirow{2}{*}{\textbf{Window}}
& \textbf{Pose}
& \textbf{Contacts}
& \multicolumn{4}{c}{\textbf{Forces}}
& \multicolumn{2}{c}{\textbf{Residual}} \\
\cmidrule(lr){2-2}
\cmidrule(lr){3-3}
\cmidrule(lr){4-7}
\cmidrule(lr){8-9}
& MPJPE [mm]$\downarrow$
& F1$\uparrow$
& MAE [N]$\downarrow$
& Angle [$^\circ$]$\downarrow$
& Corr$\uparrow$
& Share [pp]$\downarrow$
& Force [bw]$\downarrow$
& Moment [bw\,m]$\downarrow$ \\
\midrule
No mask (240-frame window)
& 58.1 & 0.940 & 86.3 & 15.5 & 0.756 & 8.2 & 0.153 & 0.040 \\
$\pm0.25\,\mathrm{s}$ (default)
& 56.6 & 0.936 & 87.5 & 15.0 & 0.729 & 8.6 & 0.155 & 0.037 \\
$\pm0.125\,\mathrm{s}$
& 56.4 & 0.933 & 88.2 & 15.1 & 0.721 & 8.7 & 0.157 & 0.038 \\
\bottomrule
\end{tabular}}
\end{table}

\subsection{Comparison with the optimization}
\label{app:vs_optim}

The optimization of \secref{sec:training_data} produces the training labels, and \tabref{tab:optim_contacts}, \tabref{tab:optim_forces} and \tabref{tab:optim_opencap} compare the two on every benchmark with the protocols of \secref{sec:experiments}.
The learned model is at least as good as its teacher almost everywhere.
On \datasetC and \wall it detects contacts with a higher recall at a similar precision, on \wall it lowers the force error on every limb and improves angle, correlation and load share, and on OpenCap it is better on every axis with either mass.
On the parkour sequences the two are within a newton of each other on average, the optimization being better on the right foot and marginally on the right hand, and the learned model on the left foot and the left hand.
The optimization keeps a smaller physical residual, since its solver minimizes exactly that residual (\appref{app:metrics_forces}). Note that these results are either on videos filtered for having good results with optimization (\datasetC) or simpler scenarios with static camera (\wall), while in the main paper we demonstrate that we beat our annotation pipeline in complex, moving camera setups.

\begin{table}[h]
\caption{\textbf{Optimization vs.\ \ours: contacts.} On \datasetC{} and \wall, with the protocol of \tabref{tab:contacts}.}
\label{tab:optim_contacts}
\centering
\small
\setlength{\tabcolsep}{4pt}
\begin{tabular}{lccc ccc}
\toprule
& \multicolumn{3}{c}{\textbf{\datasetC}}
& \multicolumn{3}{c}{\textbf{\wall}} \\
\cmidrule(lr){2-4}
\cmidrule(lr){5-7}
Method
& Prec.$\uparrow$ & Rec.$\uparrow$ & F1$\uparrow$
& Prec.$\uparrow$ & Rec.$\uparrow$ & F1$\uparrow$ \\
\midrule
Optimization
& 0.938 & 0.850 & 0.892
& \textbf{0.902} & 0.931 & 0.916 \\
\ours
& \textbf{0.948} & \textbf{0.895} & \textbf{0.921}
& 0.896 & \textbf{0.980} & \textbf{0.936} \\
\bottomrule
\end{tabular}
\end{table}

\begin{table}[h]
\caption{\textbf{Optimization vs.\ \ours: forces.} On \wall and on the parkour sequences of \citet{li2019_forces}, with the protocol of \tabref{tab:forces}. The residual of the optimization is that of its own solver, which minimizes it.}
\label{tab:optim_forces}
\centering
\small
\setlength{\tabcolsep}{2pt}
\resizebox{\linewidth}{!}{%
\begin{tabular}{@{}ll*{11}{c}@{}}
\toprule
& \multirow{2}{*}{\textbf{Method}}
& \multirow{2}{*}{\textbf{MPJPE [mm]}$\downarrow$}
& \multicolumn{5}{c}{\textbf{Force MAE [N]}$\downarrow$}
& \multicolumn{3}{c}{\textbf{Force agreement}}
& \multicolumn{2}{c}{\textbf{Residual}$\downarrow$} \\
\cmidrule(lr){4-8}
\cmidrule(lr){9-11}
\cmidrule(lr){12-13}
& &
& L.\ foot & R.\ foot & L.\ hand & R.\ hand & Avg.
& Angle [$^\circ$]$\downarrow$
& Corr$\uparrow$
& Share [pp]$\downarrow$
& Force [bw] & Moment [bw\,m] \\
\midrule
\multirow{2}{*}{\wall}
& Optimization & 57.5 & 101.8 & 101.8 & 89.2 & 84.9 & 94.5 & 14.7 & 0.654 & 10.8 & \textbf{0.087} & \textbf{0.019} \\
& \ours & \textbf{49.1} & \textbf{98.4} & \textbf{90.4} & \textbf{72.9} & \textbf{80.4} & \textbf{85.5} & \textbf{14.4} & \textbf{0.725} & \textbf{8.7} & 0.167 & 0.038 \\
\midrule
\multirow{2}{*}{Parkour}
& Optimization & \textbf{65.7} & 127.3 & \textbf{124.8} & 91.5 & \textbf{65.7} & \textbf{102.3} & 20.7 & 0.666 & 15.8 & \textbf{0.089} & \textbf{0.019} \\
& \ours & 66.3 & \textbf{117.1} & 148.4 & \textbf{77.9} & 66.3 & 102.4 & \textbf{16.7} & \textbf{0.712} & \textbf{14.6} & 0.283 & 0.044 \\
\bottomrule
\end{tabular}%
}
\end{table}

\begin{table}[h]
\caption{\textbf{Optimization vs.\ \ours: OpenCap.} Ground reaction forces with the protocol of \tabref{tab:opencap}: mean absolute error per foot in percent of body weight along the vertical, anteroposterior (AP) and mediolateral (ML) axes.}
\label{tab:optim_opencap}
\centering
\footnotesize
\setlength{\tabcolsep}{3pt}
\begin{tabular}{@{}l*{12}{c}@{}}
\toprule
\multirow{3}{*}{\textbf{Method}}
& \multicolumn{6}{c}{\textbf{OC-Walking}}
& \multicolumn{6}{c}{\textbf{OC-All}} \\
\cmidrule(lr){2-7}
\cmidrule(lr){8-13}
& \multicolumn{3}{c}{Estimated mass}
& \multicolumn{3}{c}{True mass}
& \multicolumn{3}{c}{Estimated mass}
& \multicolumn{3}{c}{True mass} \\
\cmidrule(lr){2-4}
\cmidrule(lr){5-7}
\cmidrule(lr){8-10}
\cmidrule(lr){11-13}
& Vert. & AP & ML
& Vert. & AP & ML
& Vert. & AP & ML
& Vert. & AP & ML \\
\midrule
Optimization
& 19.2 & 5.4 & 2.5
& 15.2 & 5.6 & 2.6
& 13.6 & 3.9 & 3.8
& 11.5 & 4.2 & 3.8 \\
\ours
& \textbf{11.8} & \textbf{3.0} & \textbf{1.5}
& \textbf{7.8} & \textbf{3.3} & \textbf{1.6}
& \textbf{11.6} & \textbf{3.4} & \textbf{2.3}
& \textbf{9.7} & \textbf{3.7} & \textbf{2.4} \\
\bottomrule
\end{tabular}
\end{table}

\begin{figure}[h!]
\begin{center}
\includegraphics[width=0.7\linewidth]{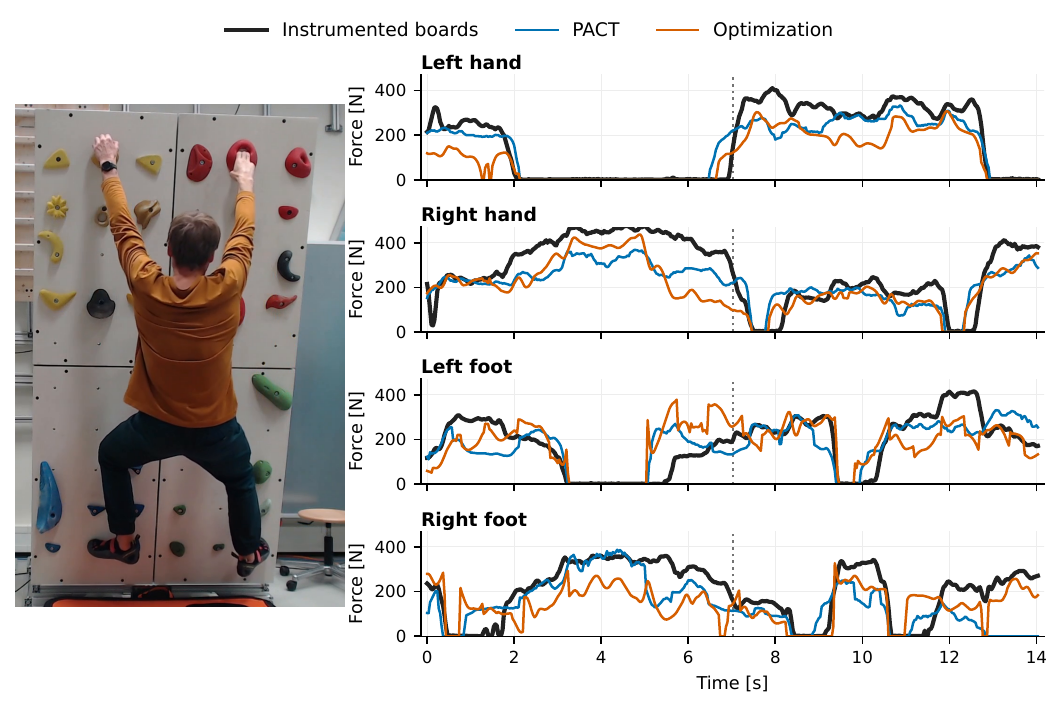}
\end{center}
\caption{\textbf{Per-limb forces on \wall.} One clip: instrumented panels (black), the optimization (orange) and the learned model (blue), in newtons with the reconstructed mass. The dotted line marks the frame shown on the left.}
\label{fig:wall_curves}
\end{figure}

\begin{figure}[h!]
\begin{center}
\includegraphics[width=0.6\linewidth]{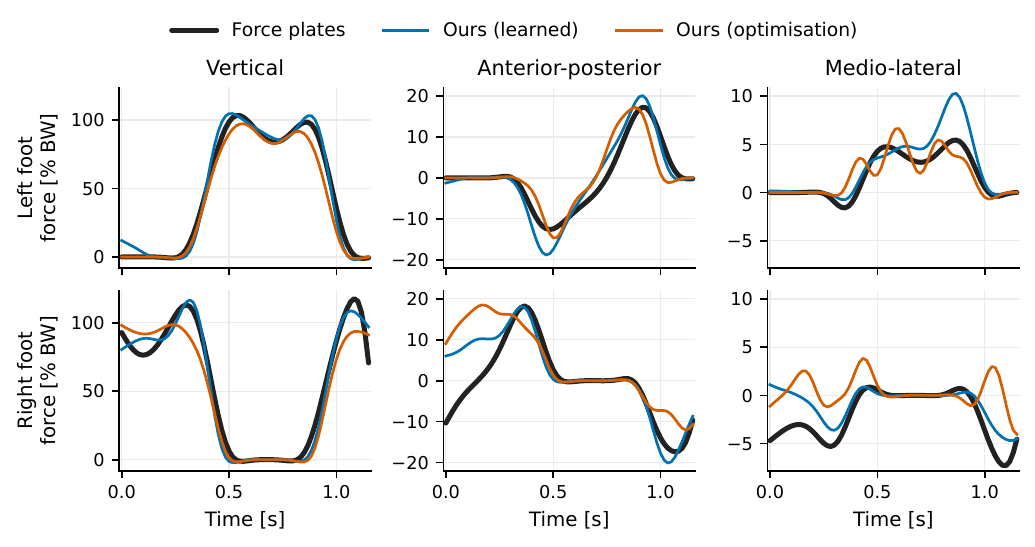}
\end{center}
\caption{\textbf{Ground reaction forces on OpenCap.} Both feet on one walking trial: force plates (black), the optimization (orange) and the learned model (blue), in percent of body weight with the measured mass.}
\label{fig:opencap_curves}
\end{figure}

\paragraph{Comparison with the force sensors.}
\Figref{fig:wall_curves} compares the per-limb forces of the optimization and of the learned model with the instrumented panels on one \wall clip, in newtons with the mass reconstructed by the pipeline, as in \appref{app:metrics_forces}, and \figref{fig:opencap_curves} compares them with the force plates on one OpenCap walking trial, in percent of body weight with the measured mass.
On the wall both estimates follow the loading and unloading of every limb, and the learned model tracks the measured magnitudes more closely, while the optimization underestimates the hands and fluctuates more from frame to frame.
On walking both recover the two peaks of the vertical force and the braking and propulsion of the anteroposterior component, with the learned model closer to the plates on both, and the mediolateral component, an order of magnitude smaller, is where both deviate most.

\subsection{More qualitative results}
\label{app:qualitative}

\begin{figure}[p]
\begin{center}
\includegraphics[width=1.0\linewidth]{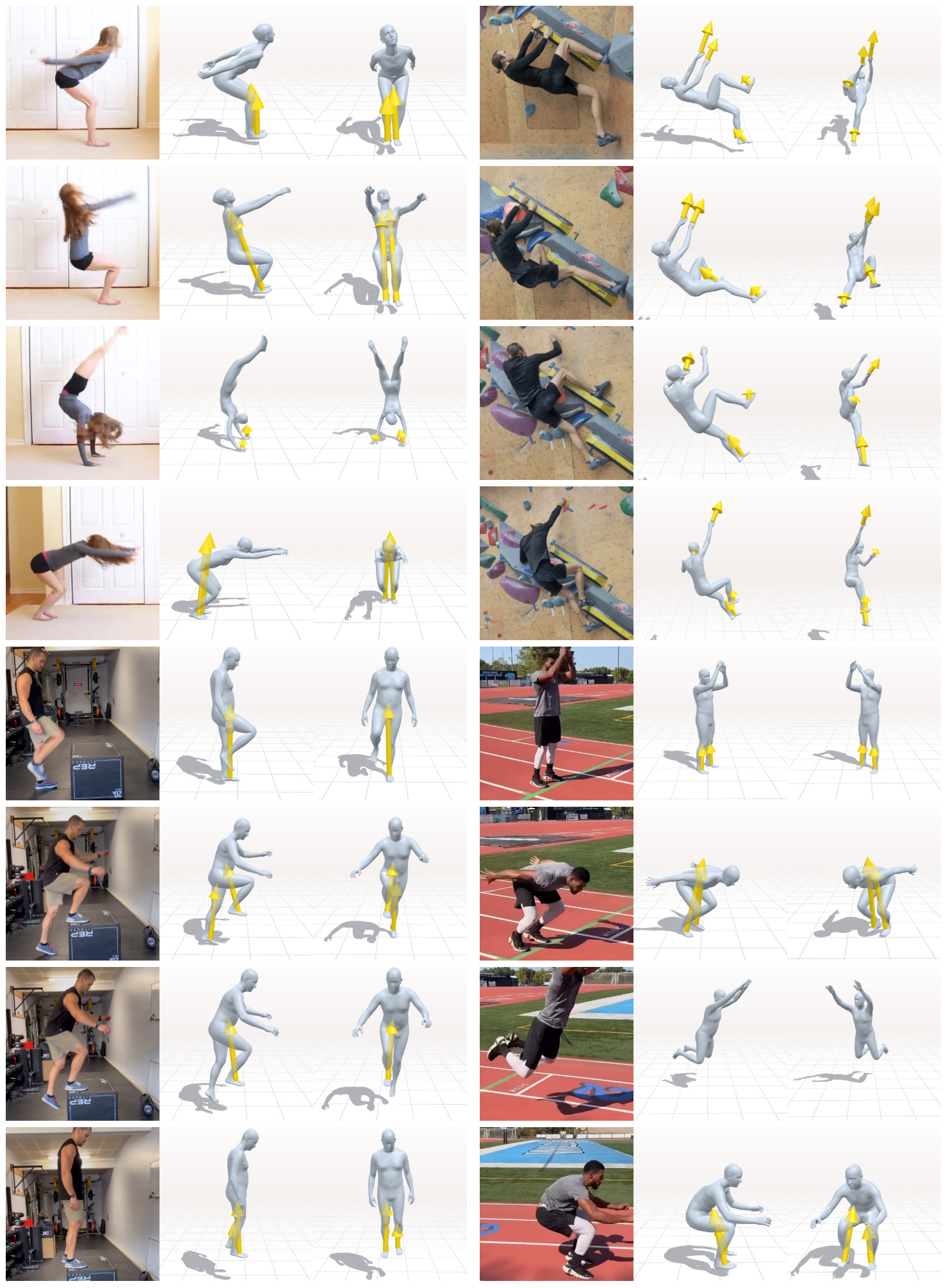}
\end{center}
\caption{\textbf{Qualitative results in 3D.} Input frame and two rendered views of the estimated body with the predicted contact forces.}
\label{fig:qual_3d}
\end{figure}

\begin{figure}[p]
\begin{center}
\includegraphics[width=1.0\linewidth]{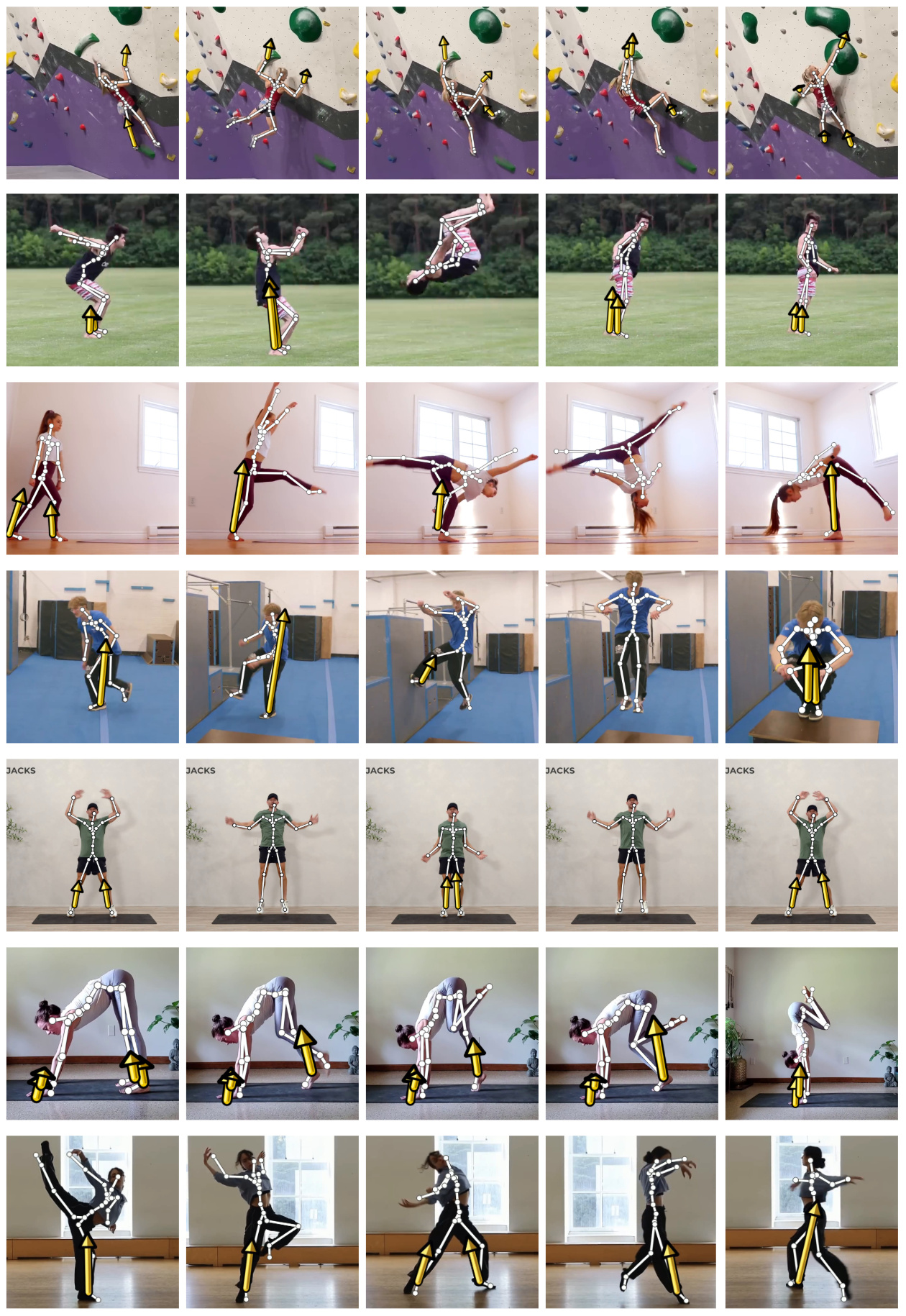}
\end{center}
\caption{\textbf{Qualitative results over time.} Consecutive frames of one clip per row with the estimated poses and the predicted contact forces.}
\label{fig:qual_motion}
\end{figure}

\Figref{fig:qual_3d} shows single frames of in-the-wild videos with the estimated body and forces rendered from the camera and a second viewpoint, and \figref{fig:qual_motion} consecutive frames of one clip per row, with poses in white and forces in yellow.

\end{document}